\documentclass[journal]{IEEEtran}
\usepackage{amsmath,amsfonts}
\usepackage{algorithmic}
\usepackage{algorithm}
\usepackage{array}
\usepackage[
colorlinks=true,
linkcolor=blue,
urlcolor=blue,
citecolor=blue
]{hyperref}
\usepackage{textcomp}
\usepackage{stfloats}
\usepackage{url}
\usepackage{verbatim}
\usepackage{graphicx}
\usepackage{capt-of}
\usepackage{cite}
\usepackage{multirow}
\usepackage{multicol}
\usepackage{graphicx}
\usepackage[caption=false]{subfig}
\usepackage[table]{xcolor}
\usepackage{colortbl}
\usepackage{makecell}
\usepackage{booktabs}
\usepackage{amsmath}
\usepackage{amssymb}
\usepackage{pdfpages}
\usepackage{newfloat}
\usepackage{picins}
\usepackage{listings}
\usepackage{float}
\usepackage{bm}
\usepackage{orcidlink} %调包
\hypersetup{hidelinks} % hide the hyperlink box
\newcommand\blfootnote[1]{%
  \begingroup
  \renewcommand\thefootnote{}\footnote{#1}%
  \addtocounter{footnote}{-1}%
  \endgroup
}
\begin{document}

\title{IV-tuning: Parameter-Efficient Transfer Learning for \\ Infrared-Visible Tasks}

% \author{IEEE Publication Technology,~\IEEEmembership{Staff,~IEEE,}
%         % <-this % stops a space
% \thanks{This paper was produced by the IEEE Publication Technology Group. They are in Piscataway, NJ.}% <-this % stops a space
% \thanks{Manuscript received April 19, 2021; revised August 16, 2021.}}
% \author{
% Yaming Zhang{$^{1}$} ~ Chenqiang Gao{$^{2*}$}~ Fangcen Liu{$^{1}$} ~ Junjie Guo{$^{1}$} ~ \\Lan Wang{$^{3}$}~ Xinggan Peng{$^{4}$}~ Deyu Meng{$^{5}$}~ \\
% \normalsize
% $^{1}$\	Chongqing University of Posts and Telecommunications ~~$^{2}$\,Shenzhen Campus of Sun Yat-sen University ~~ \\
% $^{3}$\,Michigan State University ~~
% \normalsize
% $^{4}$\,Nanyang Technological University ~~
% $^{5}$\,Xi’an Jiaotong University\\
% \normalsize
% {\tt\small \{YummyZhang1989,liufc67,gjj893866738\}@gmail.com},
% {\tt\small gaochq6@mail.sysu.edu.cn}, \\
% {\tt\small wanglan3@msu.edu}, 
% {\tt\small xinggan001@entu.edu.sg}, 
% {\tt\small dymeng@mail.xjtu.edu.cn}
% }
\author{
    Yaming~Zhang$^{\orcidlink{0009-0006-5333-7258}}$,
    Chenqiang~Gao$^{\orcidlink{0000-0003-4174-4148}}$,
    Fangcen~Liu$^{\orcidlink{0000-0002-1845-9907}}$,
    Junjie~Guo$^{\orcidlink{0009-0005-8957-8419}}$,
    Lan~Wang,
    Xinggan~Peng$^{\orcidlink{0000-0002-1087-8576}}$,
    and~Deyu~Meng$^{\orcidlink{0000-0002-1294-8283}}$
}

% The paper headers
% \markboth{Journal of \LaTeX\ Class Files,~Vol.~14, No.~8, August~2021}%
% {Shell \MakeLowercase{\textit{et al.}}: A Sample Article Using IEEEtran.cls for IEEE Journals}
\markboth{IEEE Transactions on Circuits and Systems for Video Technology,~Vol.~XX, No.~XX, Month~Year}{Author \MakeLowercase{\textit{et al.}}: Title}
% \IEEEpubid{0000--0000/00\$00.00~\copyright~2021 IEEE}
% Remember, if you use this you must call \IEEEpubidadjcol in the second
% column for its text to clear the IEEEpubid mark.

\twocolumn[{%
    \renewcommand\twocolumn[1][]{#1}%
    \maketitle
    \setlength{\abovecaptionskip}{0.1cm}
    \setlength{\belowcaptionskip}{0.1cm}
    \begin{center}
        \centering
        \includegraphics[width=1\textwidth]{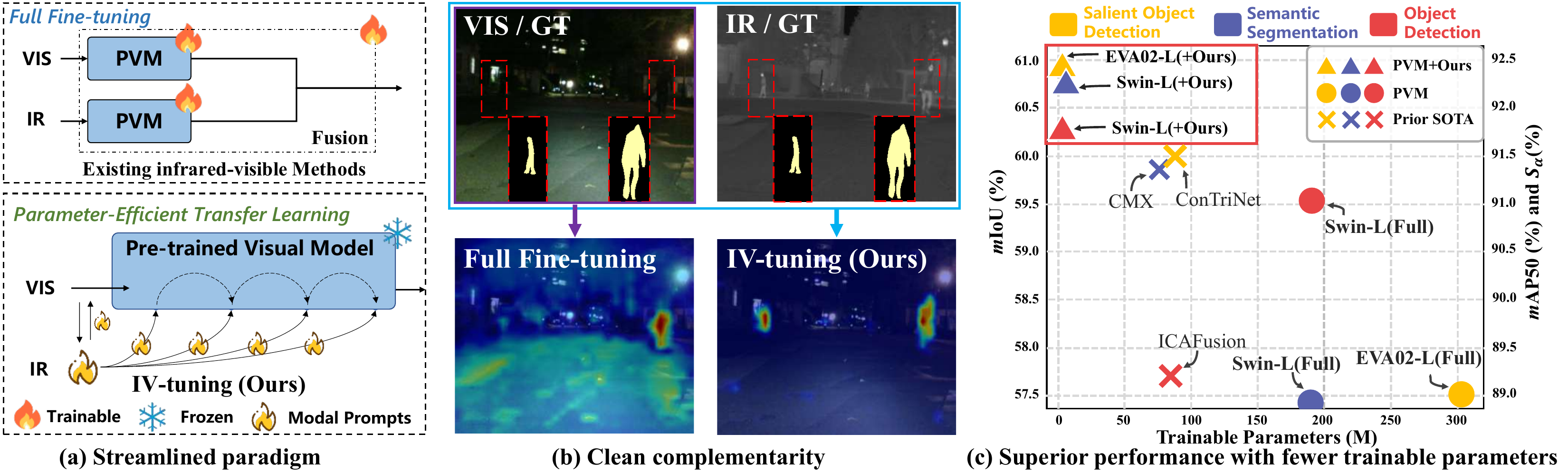}
        \captionof{figure}{(a) top: Existing infrared-visible (IR-VIS) methods typically extend Pre-trained Visual Models (PVMs) into a dual-branch network and perform full fine-tuning. 
        (a) bottom: We propose a streamlined paradigm where the utilization of infrared can be greatly simplified. 
        (b) Fully fine-tuning the PVM leads to overfitting on background regions (as evidenced in Fig. \ref{fig:pca}), whereas our method alleviates overfitting and learns the complementarity between modalities effectively.
        (c) Our model demonstrates superior performance and generality with fewer trainable parameters.}
        
        \label{head} % label 放在 captionof 之后
    \end{center}
}]
\global\let\thanks\IEEEthanks
\blfootnote{\hspace*{0.5em}(Corresponding author: Chenqiang Gao.)
\\ % 换行，并增加 2pt 的行间距，避免太挤
\hspace*{1.5em}Yaming Zhang, Fangcen Liu, and Junjie Guo are with the School of Communications and Information Engineering, Chongqing University of Posts and Telecommunications, Chongqing 400065, China (e-mail: \{YummyZhang1989, liufc67, gjj893866738\}@gmail.com).
\\
\hspace*{1.5em}Chenqiang Gao is with the School of Intelligent Systems Engineering, Shenzhen Campus of Sun Yat-sen University, Shenzhen 518107, China (e-mail: gaochq6@mail.sysu.edu.cn).
\\
\hspace*{1.5em}Lan Wang, Independent Researcher (e-mail: wanglan.research@gmail.com).
\\
\hspace*{1.5em}Xinggan Peng is with Nanyang Technological University, Singapore 639798 (e-mail: xinggan001@ntu.edu.sg).
\\
\hspace*{1.5em}Deyu Meng is with the School of Mathematics and Statistics, Xi’an Jiaotong University, Xi’an 710049, China (e-mail: dymeng@mail.xjtu.edu.cn).
\\
\hspace*{1.5em}Code is available at \url{https://github.com/Yummy198913/IV-tuning}.
}

\begin{abstract}
Existing infrared and visible (IR-VIS) methods inherit the general representations of Pre-trained Visual Models (PVMs) to facilitate complementary learning.
However, our analysis indicates that under the full fine-tuning paradigm, the feature space becomes highly constrained and low-ranked, which has been proven to seriously impair generalization.
One remedy is to freeze the parameters, which preserves pretrained knowledge and helps maintain feature diversity.
To this end, we propose IV-tuning, to parameter-efficiently harness PVMs for various IR-VIS downstream tasks, including salient object detection, semantic segmentation, and object detection.
Extensive experiments across various settings demonstrate that IV-tuning outperforms previous state-of-the-art methods, and exhibits superior generalization and scalability.
Remarkably, with only a single backbone, IV-tuning effectively facilitates the complementary learning of infrared and visible modalities with merely 3\% trainable backbone parameters, and achieves superior computational efficiency compared to conventional IR-VIS paradigms.
\end{abstract}

\section{Introduction}
\label{sec:intro}
\begin{figure*}[ht]
  \centering
  \includegraphics[width=1\linewidth]{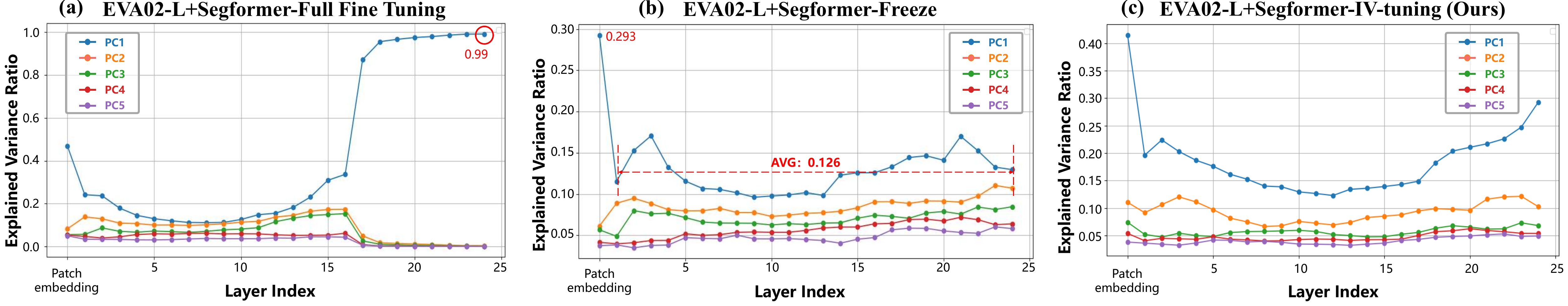}
  \caption{Observation of effective information contained in the feature space via Principal Component Analysis (PCA). 
  We apply PCA for dimension reduction and visualize the explainable variance ratio of principal components with high contributions.
  We track the first five principal components across layers of the EVA02-L \cite{fang2024eva} + Segformer \cite{xie2021segformer} model and illustrate how the feature space evolves as depth increases.
  We show that: (a) fully fine-tuned model quickly converges to a highly constrained and low-ranked subspace in higher layers, thereby sacrificing generalization ability.
  (b) frozen model maintains feature diversity but struggles to extract task-specific discriminative information. 
  In contrast, our model (c) not only preserves the feature diversity of subspaces but also effectively excavates task-specific discriminative information.
  }
  \label{fig:pca}
\end{figure*}
\begin{figure}[t]
  \centering
  \includegraphics[width=1\linewidth]{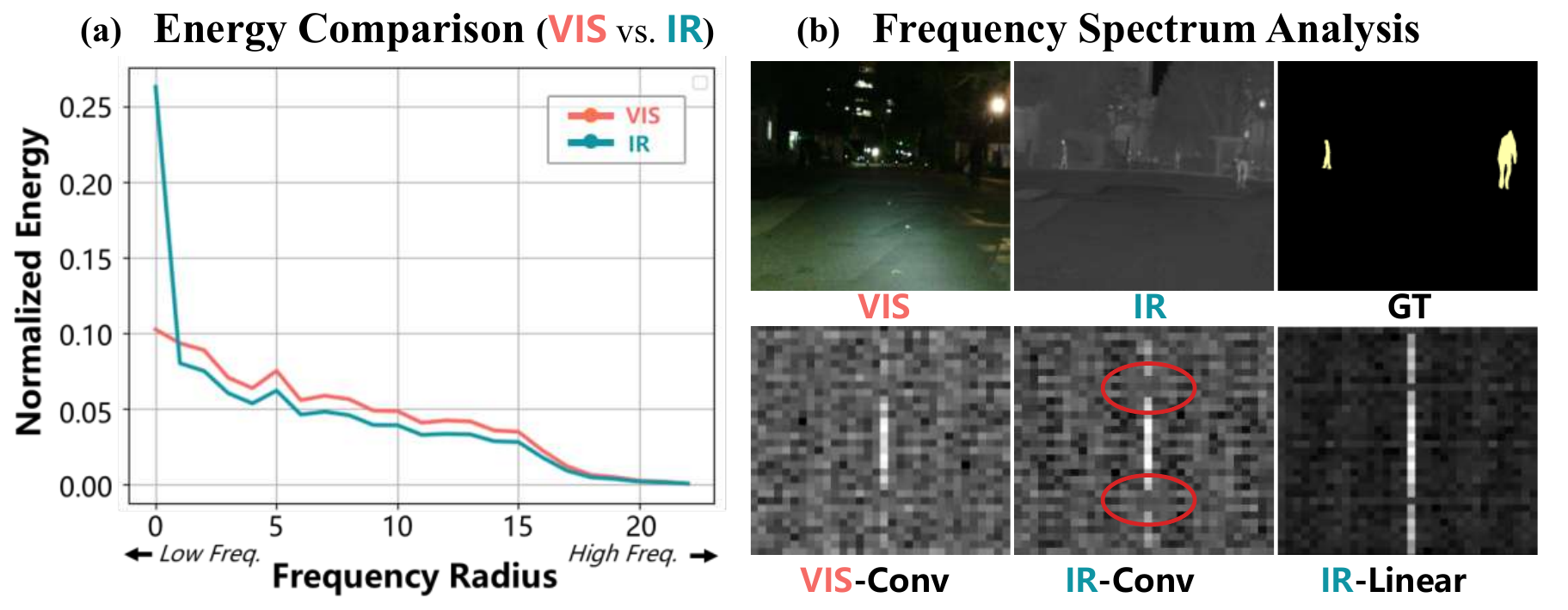}
  % \vspace{-13pt}
  \caption{Analysis of energy distribution between infrared and visible modalities, and frequency spectral patterns under varying conditions, where the center represents low frequency and the corners depict high frequency. 
  We calculate the statistical average distribution of energy on the MFNet \cite{ha2017mfnet} training set.
  We show that (1) compared with visible image, infrared modality exhibits a stronger low-frequency response, while their mid-to-high-frequency share certain similarities with those of visible modality.
  (2) convolutional layer enhances high‑frequency details—boosting texture and edge information in visible images.
  However, it causes the loss of low-frequency in the infrared modality, which are critical for complementary learning.
  In contrast, simple linear projection can effectively capture low-frequency signals in the infrared modality, which inspires the motivation behind our design.
  }
  \label{fig:energy}
\end{figure}
\IEEEPARstart{I}{n} Computer Vision, various tasks, such as salient object detection, semantic segmentation and object detection, have predominantly relied on the visible modality, and numerous visible-based methods \cite{hou2024salience,wei2024stronger,zong2023detrs,wang2023yolov7,zhang2023mp, zhou2021ecffnet, song2022multiple} have emerged over the past decades.
However, the performance of visible-based methods degrades in challenging scenarios (e.g., nighttime, fog, rain) due to the inherent limitations of visible imaging principles. 
Therefore, leveraging the infrared and visible images with complementary information to improve performance has increasingly drawn attention.
Various methods \cite{gu2025discovering,huang2025t,guo2024damsdet,zhao2024equivariant,shen2024icafusion,li2024residual,park2023cross,tang2024divide} have been widely explored for the infrared-visible tasks with classic backbones, such as ResNet \cite{he2016deep}, VGGNet \cite{simonyan2014very} and CSPDarkNet \cite{bochkovskiy2020yolov4}.

With the introduction of the Transformer \cite{vaswani2017attention}, transformer-based Pre-trained Visual Models (PVMs), such as ViT \cite{dosovitskiy2020image}, Swin Transformer \cite{liu2021swin}, MAE \cite{he2017mask}, EVA02 \cite{fang2024eva} and DINOv3 \cite{simeoni2025dinov3} have demonstrated powerful generalization across various downstream tasks.
Some infrared-visible methods \cite{shen2024icafusion,zhao2023cddfuse,liu2021swinnet} have begun to leverage the general representations of PVMs to improve performance.
As illustrated in Fig. \ref{head} (a) top, these methods routinely design task-specific fusion strategies and fully fine-tune a dual-branch model to learn the complementarity of the two modalities.
Albeit effective, the resulting models are costly to train, hard to scale, and prone to overfitting---especially given the small datasets typical of the IR–VIS community \cite{liu2022target,ha2017mfnet}.

From a modern viewpoint, the scaling law \cite{kaplan2020scaling} suggests that on a smaller fixed dataset, performance ceases to improve as model parameters increase, leading to overfitting.
To investigate this, we analyze the behavior of feature spaces with Principal Component Analysis (PCA) under full fine-tuning and freezing paradigms.
As the network depth increases, the fully fine-tuned model initially captures diverse signals but quickly falls into a highly constrained and low-ranked\footnote{The ``rank" here means the number of significant principal components.} subspace, where a single principal component can capture almost all effective information\footnote{In PCA, it refers to variance captured by principal components. Larger eigenvalues indicate components explaining more variance and contributing more to data representation.}.
This behavior aligns with early information-bottleneck studies of neural networks \cite{shwartzziv2017openingblackboxdeep,gunasekar2017implicit} that overfitting is largely due to the compression phase and low-ranked feature spaces hinder generalization by memorizing trivial patterns.

One potential remedy to address the overfitting problem is freezing the parameters.
As shown in Fig. \ref{fig:pca} (b), the frozen PVM rapidly expands the input feature space after the first layer, leading to a more expressive representation, but fails to capture task-specific discriminative information.
Therefore, Parameter-Efficient Transfer Learning (PETL) introduces lightweight modules on top of frozen models, such as inserting adapters \cite{chen2022adaptformer} or learnable prompts \cite{zhu2023visual,jia2022visual}, offering a favorable adaptation pathway.

A recent work \cite{yuan2024unirgb} in the IR-VIS community introduced multi-scale and attention modules into the frozen backbone, which effectively reduce the computational overhead of IR-VIS tasks while achieving considerable performance.
However, like most previous IR-VIS methods, it fails to account for the intrinsic heterogeneity between infrared and visible modalities fully.
Hence, how to effectively and efficiently harness the infrared modality remains an open question and warrants further exploration.

Visually, infrared images exhibit distinct thermal radiation between objects and background.
Such coarse thermal structures correspond to low-frequency variations in the frequency domain.
To quantify this observation, we analyze the energy distributions\footnote{The ``energy" here is defined as the sum of Fourier magnitudes within each radial frequency band, normalized to form a distribution. It reflects the proportion of information across different frequency components.} of both modalities.
As shown in Fig. \ref{fig:energy} (a), the infrared modality has significantly stronger energy responses in the low-frequency region, while its mid-to-high-frequency energy distribution shows a notable similarity to that of the visible modality.
This indicates that the low-frequency components of infrared need to be explicitly preserved, while common convolution operations (e.g., 3$\times$3 convolution), due to their limited receptive fields, weaken the low-frequency signals, as shown in Fig. \ref{fig:energy} (b).
In contrast, linear projection inherently applies a global transformation across the feature space, which tends to preserve low-frequency information \cite{rahaman2019spectral,xu2019frequency}.
This minimalistic yet effective characteristic renders linear projection highly suitable for parameter-efficient adaptation, allowing smooth integration of infrared priors into frozen PVMs.

Based on above analysis, we present IV-tuning, a parameter-efficient method to harness the PVMs for IR-VIS tasks, including salient object detection, semantic segmentation and object detection.
As illustrated in Fig. \ref{head} (a) bottom, instead of adding an extra backbone, IV-tuning introduces cascade modal prompts to the frozen PVMs, which inherits the general representations of PVMs to the maximum extent.
Specifically, we propose the Modality-aware Prompter (MP) to effectively learn the inter-modal complementarities, which applies modality‑specific operations to infrared and visible features, generating modal prompts that capture and fuse their complementary strengths.

Furthermore, we observe a sharp drop in feature rank after the first layer of PVM, as shown in Fig. \ref{fig:pca} (b) (0.293 vs. 0.126).
This quantitative change indicates a phase transition in the feature space: from a concentrated, low-intrinsic-dimension subspace in shallow layers to a diverse, high-entropy manifold in deep layers, where we argue that a uniform fusion mechanism is suboptimal.
To this end, we propose two versions of fusion strategies (denoted as $\alpha$-fusion and $\beta$-fusion) to address this discrepancy, and show that high‑dimensional fusion ($\beta$) better adapts to the diverse feature space.
The main contributions of our work can be summarized below:
\begin{itemize}
    \item
    New perspective to analyze the overfitting risk in IR-VIS tasks: 
    We point out that with the widespread application of pre-trained visual models nowadays, there is an urgent need to analyze the performance of these data-driven models under the full fine-tuning paradigm. Our PCA indicates that PVM under the full fine-tuning paradigm highly constrains model’s expressivity, thereby limiting its generalization ability for IR-VIS tasks.
    \item
    Key insights into the complementary information between infrared and visible modalities: 
    We reveal, via energy distribution and frequency spectra, that the key differences between two modalities lie in low-frequency components, which need to be explicitly preserved. Moreover, the convolution degrades such signals while linear layers preserve them, which motivates our design.
    \item
    Novel Approach via Parameter-Efficient Transfer Learning: 
    We propose a general and efficient framework, \textit{IV-tuning}, with two designs: (1) freezing the backbone to preserve pre-trained knowledge while inserting cascade MP blocks to effectively learn the task-specific discriminative information. (2) a differential design tailored to the intrinsic disparities between two modalities. Moreover, we propose two fusion strategies $\alpha$-fusion for low-ranked space and $\beta$-fusion for diverse space, thereby ensuring tailored adaptations.
    \item
    Extensive experiments to demonstrate the effectiveness:
    By training less than 3\% of the backbone parameters, IV-tuning achieves superior performance across 2 pre-trained models, 5 datasets of 3 tasks, including IR-VIS salient object detection, semantic segmentation and object detection, outperforming the full fine-tuning paradigm and previous state-of-the-art methods, as shown in Fig. \ref{head} (c).
\end{itemize}

The remainder of this paper is organized as follows: Section \ref{sec:related} discusses related works.
Section \ref{sec:methodology} presents the proposed method.
Experimental results and analysis are provided in Section \ref{sec:experiment}.
Finally, the conclusion is drawn in Section \ref{conclusion}.

\begin{figure*}[ht]
  \centering
  \includegraphics[width=1.0\linewidth]{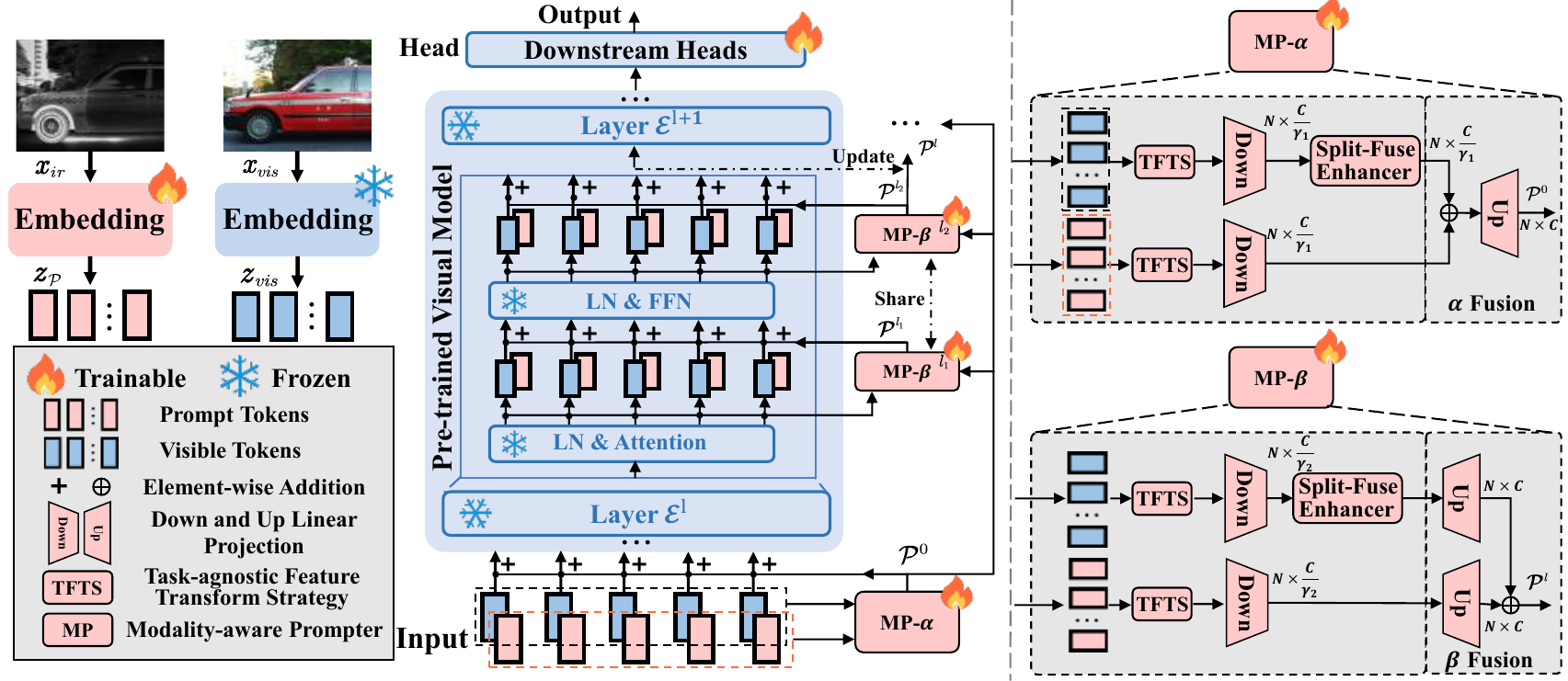}
  % \vspace{-15pt}
  \caption{The overview of the proposed IV-tuning. IV-tuning freezes the $L$-layer transformer-based backbone and only fine-tunes a select few modules to learn effective visual prompts. The initial prompt $\bm{\mathcal{P}}^{0}$ is generated by MP-$\alpha$ block and sent to each MP-$\beta$ block, with each encoder's output updating the initial prompt $\bm{\mathcal{P}}^{0}$.}
  \label{fig:overview}
  % \vspace{-10pt}
\end{figure*}
\section{Related Work}
\label{sec:related}
\subsection{Infrared-visible Tasks}
Existing IR–VIS methods for tasks, such as semantic segmentation, object detection, and salient object detection fall into two categories:
The first adopts the image-level fusion paradigm. It first fuses infrared and visible images to generate a fused image, then feeds it into a single-branch network.
The second category represents the end-to-end fusion paradigm, adopting a dual-branch architecture without independent pre-fusion. IR and VIS images are processed by separate backbones, with feature fusion at encoder or decoder layers.

A key limitation of most methods is over-reliance on dual-branch architectures. To balance computation, they usually adopt lightweight, even outdated backbones (e.g., MobileNet \cite{howard2017mobilenets,sandler2018mobilenetv2}, ResNet \cite{he2016deep}), with a focus on complex decoder designs to compensate for weak backbones and alleviate modal differences, which not only limits the use of advanced PVMs but also reduces method generality—complex decoders are task-specific and cannot be migrated across tasks.
A recent work \cite{yuan2024unirgb} introduced PETL to IR-VIS tasks yet failed to fully consider IR-VIS modal differences. In conclusion, how to effectively utilize infrared information and modern PVMs remains an open question warranting further exploration.

\subsection{Pre-trained Visual Models}
Pre-trained visual models (PVMs) have long served as fundamental backbones for diverse vision tasks, evolving from convolutional architectures to modern Transformer-based foundations.
% Early dominant frameworks were built upon Convolutional Neural Networks (CNNs), typically pre-trained on the ImageNet-1K benchmark \cite{deng2009imagenet}, which provides large-scale annotated data for learning general low-to-high level visual features.
Representative Convolutional Neural Networks such as AlexNet \cite{krizhevsky2012imagenet}, VGG \cite{simonyan2014very}, and ResNet \cite{he2016deep} advanced feature extraction via deepened structures and residual connections.
With the rise of the Transformer architecture \cite{vaswani2017attention}, pre-trained visual models have entered a new era. Vision Transformer (ViT) \cite{dosovitskiy2020image} pioneered the fully attentional visual paradigm, while Swin Transformer \cite{liu2021swin} further improved efficiency and adaptability through hierarchical structure and window-based attention. Driven by self-supervised learning, advanced vision foundation models have emerged, such as MAE \cite{he2022masked}, utilizing masked image modeling for learning latent image representations; CLIP, which learns high quality visual representation by exploring contrastive learning with large-scale image text pairs; EVA02 \cite{fang2024eva}, which integrates Masked Image Modeling pre-training with CLIP’s vision features, and DINO series \cite{oquab2023dinov2,simeoni2025dinov3}, which is pretrained on extensive, curated datasets without explicit supervision. These models have shown superior performance on standard vision benchmarks.
 
However, such powerful pre-trained models face critical limitations when deployed for infrared-visible tasks. The mainstream PVMs are pre-trained exclusively on visible data, lacking inherent adaptation to the distinct characteristics of infrared modality. Meanwhile, over-parameterized PVMs suffer overfitting risks when fully fine-tuned on small-scale downstream datasets, leading to catastrophic forgetting of pre-trained knowledge and degraded generalization. Moreover, the ever-increasing model scale significantly hinders the practicality of dual-branch fusion structures, introducing heavy computational overhead, larger memory footprint, and higher deployment complexity for IR-VIS scenarios.
Given that the use of PVMs has become a widespread trend in the IR-VIS community, addressing their potential overfitting risks and efficiently extending them to IR-VIS tasks has become an urgent need.
% As the transformer \cite{vaswani2017attention} rises, larger visual models (e.g., ViT \cite{dosovitskiy2020image}, Swin Transformer \cite{liu2021swin}, MAE\cite{he2017mask} and EVA02\cite{fang2024eva}) are pre-trained on larger datasets, demonstrating superior scalability, long-range dependency modeling, and representation learning capabilities.
% However, the huge appetite of these data-driven models increases the potential risk of overfitting, especially on small-scale datasets \cite{kaplan2020scaling}.
% Moreover, the growing size of PVMs also hinders the scalability of dual-branch fusion architectures, raising the cost and complexity of adapting them to IR-VIS tasks.
% Given that the use of PVMs has become a widespread trend in the IR-VIS community, addressing their potential overfitting risks and efficiently extending them to IR-VIS tasks has become an urgent need.

\subsection{Visual Parameter-Efficient Transfer Learning}
The advent of large pre‑trained visual models (PVMs) has fueled growing interest in Parameter‑Efficient Transfer Learning (PETL), a paradigm that freezes the PVMs and only tunes a small subset of task-specific parameters. PETL effectively addresses the drawbacks of full fine-tuning, such as excessive GPU memory consumption, high training costs, and overfitting risks on small-scale datasets, making it widely adopted in various downstream tasks. Two popular PETL paradigms are already well-established:

The first is Prompt‑tuning \cite{jia2022visual,pei2024sa2vp,zhu2023visual}, which introduces learnable prompt tokens or embeddings into the input space of the pre-trained backbone. These prompts can guide the backbone to extract task-relevant features without modifying its original parameters. In visual tasks, prompt-tuning typically embeds visual prompts into the input space, enabling flexible adaptation to downstream tasks while retaining the backbone’s powerful representation capabilities.
The second is Adapter‑tuning \cite{chen2022adaptformer,li2024adapter,yin20231}, which injects lightweight bottleneck modules into the transformer encoder layers of the backbone. These adapter modules have a small number of parameters and are trained independently, allowing them to capture task-specific information without disrupting the pre-trained feature distribution, thus achieving efficient transfer learning.

All the aforementioned PETL methods can match or even exceed full fine-tuning performance with fewer trainable parameters. However, most existing works focus on single-task or single-modality scenarios \cite{wei2024stronger,pei2024sa2vp,li2024adapter,zhu2023visual}, failing to consider the inherent differences and complementary characteristics between different modalities. In this paper, we aim to integrate the distinct properties of infrared and visible data and propose a general PETL method tailored for IR-VIS multi-modal tasks.

\section{Methodology}
\label{sec:methodology}
\subsection{Overall Architecture}
Compared to visible tasks, the infrared-visible tasks introduce an extra infrared input $\bm{x}_{ir}$, which is temporally synchronized and spatially aligned with the visible input $\bm{x}_{vis}$.
These tasks aim to learn the function $F:\{\bm{x}_{vis}, \bm{x}_{ir}\}\rightarrow\bm{p}$ to obtain the result $\bm{p}$.
Typically, the function $F$ can be decomposed into $\phi\circ f$, where $f$ represents the feature extraction and interaction function, and $\phi$ is a task-oriented head producing the final prediction $\bm{p}$.
Hereby, the $f$ is a transformer-based Pre-trained Visual Models (PVMs) in our case, which includes the norm layer $LN(\cdot)$, Multi-head Self-Attention $Attn(\cdot)$ and the Feed-Forward Network $FFN(\cdot)$.

To inject infrared knowledge, we first apply an independent patch embedding layer to $\bm{x}_{ir}$, yielding prompt tokens $\bm{z}_{\mathcal{P}} \in \mathbb{R}^{N \times C}$ that share dimensionality $N$ (token length) and $C$ (embedding dimension) with the visible tokens $\bm{z}_{vis} \in \mathbb{R}^{N \times C}$, as shown in Fig. \ref{fig:overview}.
Then, these tokens $\{\bm{z}^{0}_{vis}, \bm{z}_{\mathcal{P}}\}$ are fed into an MP-$\alpha$ block to generate the initial modal prompt $\bm{\mathcal{P}}^{0}$, which is then sent into cascade MP-$\beta$ blocks.

Within each encoder layer $\mathcal{E}^{l}$ ($l=1,2,3,\ldots,L$), we insert two weight-sharing MP-$\beta$ blocks after the $Attn$ layer and $FFN$ layer, respectively, to progressively refine the backbone feature.
Denoting the input to layer $\mathcal{E}^{l}$ as $\bm{z}^{l-1}$, the process of the $l$-th encoder layer can be formulated as:
\begin{align}
\bm{z}^{l} = FFN(Attn(\bm{z}^{l-1}) + \bm{\mathcal{P}}^{l_{1}}) + \bm{\mathcal{P}}^{l_{2}},
\end{align}
where the $\bm{\mathcal{P}}^{l_{1}}$ and $\bm{\mathcal{P}}^{l_{2}}$ are the output of the block MP-$\beta^{l_{1}}$ and MP-$\beta^{l_{2}}$, respectively, and we omit the residual connection for brevity.
The output of each encoder layer $\bm{z}^{l}$ updates the initial modal prompt for next layer.
After $L$ encoder layers, the final token sequence $\bm{z}^{L}$ is passed to the decoder $\phi$ to generate the prediction: $\bm{p} = \phi(\bm{z}^{L})$.
In this manner, the MP-$\alpha$ and MP-$\beta$ blocks capture the inter-modal complementarity at every semantic level, enabling robust adaptation of PVMs to the downstream task.

\begin{figure}[t]
  \centering
  \includegraphics[width=1\linewidth]{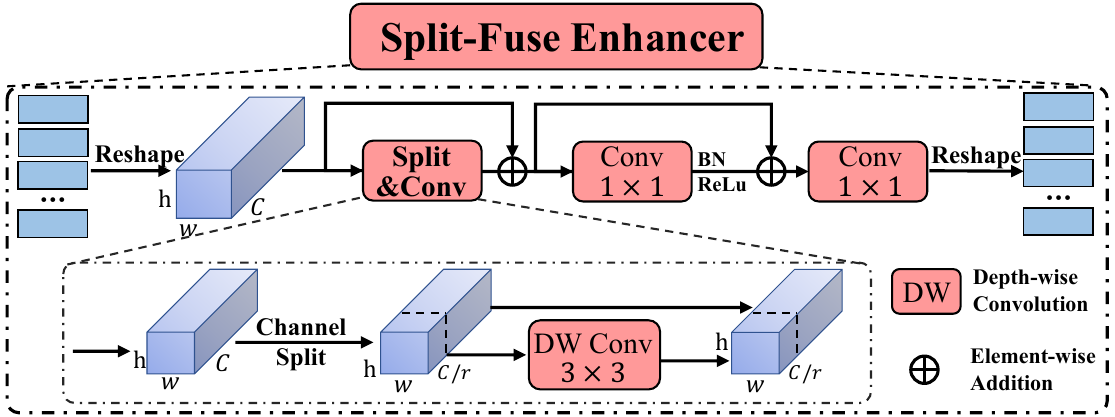}
  \caption{Detailed design of the Split-Fuse Enhancer. The input tokens are reshaped to feature maps, then a depth-wise $3\times3$ convolution is applied to selected channels with residual connection, followed by two $1\times1$ convolutional layers with Batch Normalization (BN) and ReLU activation in between, where the output maintains the same dimension as the input.
  We set the split ratio $\frac{1}{r}$ as $\frac{1}{4}$ by default.}
  \label{fig:ho}
  \vspace{-10pt}
\end{figure}
\subsection{Modality-aware Prompter}
In light of our PCA analysis shown in Fig. \ref{fig:pca}, we propose the Modality-aware Prompter (comprising MP-$\alpha$ and MP-$\beta$), which projects features into a low-ranked subspace to extract the principal components and applies modality-specific operations to leverage their complementary strengths.
As illustrated in Fig. \ref{fig:overview}, the MP-$\alpha$ and MP-$\beta$ share the similar structure but differ in their fusion strategies.
For simplicity, we first describe the identical forward process of the two blocks, and later elaborate on the differences in their fusion method.

\textit{1) Task-agnostic Feature Transform Strategy:}
pre-trained visual models have demonstrated their advantages in generating visual representations. 
However, these general representations are often task-agnostic, thus requiring fine-tuning for downstream tasks. 
Thus, we propose a task-agnostic feature transformation strategy to enable a simple and efficient scalable adaptation.
Formally, given visible tokens and prompt tokens $\{\bm{z}_{vis}, \bm{z}_{\mathcal{P}}\}$, two learnable matrices $\bm{\omega}$ and $\bm{\phi}$ are introduced for distribution normalization and channel-wise recalibration:
\begin{align}
    \bm{z}^{\prime}_{vis} =LN_{1}(\bm{z}_{vis}) \odot \bm{\omega_{1}}+\bm{\phi_{1}}, \\
    \bm{z}^{\prime}_{\mathcal{P}} =LN_{2}(\bm{z}_{\mathcal{P}}) \odot \bm{\omega_{2}}+\bm{\phi_{2}},
\end{align}
where $\odot$ denotes the element-wise multiplication, and $\bm{z}^{\prime}_{vis}$ and $\bm{z}^{\prime}_{\mathcal{P}}$ are scaled features.
This strategy enables a flexible scaling over features of different modalities with negligible parameter overhead.
Then, we project each flow into a latent space by:
\begin{align}
   & \bm{\mathcal{M}}_{vis}=s_{1}(\bm{z}^{\prime}_{vis}), \\
   & \bm{\mathcal{M}}_{\mathcal{P}}=s_{2}(\bm{z}^{\prime}_{\mathcal{P}}),
\end{align}
where $\bm{\mathcal{M}}_{vis}\in\mathbb{R}^{N \times \frac{C}{\gamma}}$ and $\bm{\mathcal{M}}_{\mathcal{P}}\in\mathbb{R}^{N \times \frac{C}{\gamma}}$ are latent representations, and $s_{1}(\cdot)$ and $s_{2}(\cdot)$ are linear projection layers.

% \textit{2) Modality-aware Process:}
% To effectively capture the complementary information, we design modality-aware processing branches tailored for different modalities.
% To enhance task-relevant signals in visible flow, we propose the Split-Fuse Enhancer.
% Since the PCA reveals certain channel redundancy of features, we split $\bm{\mathcal{M}}_{vis}$ into two parts: one processed by a $3 \times 3$ depth-wise convolution and concatenated with the other channels via a residual connection, as shown in Fig. \ref{fig:ho}.
% Then, two $1 \times 1$ convolutions are applied to aggregate cross-channel information to produce the enhanced representation $\bm{\mathcal{M}}^{e}_{vis}\in\mathbb{R}^{N \times \frac{C}{\gamma}}$.
% Meanwhile, the infrared representation $\bm{\mathcal{M}}_{\mathcal{P}}$ is unaltered to retain its focused semantic cues.
\textit{2) Modality-aware Process:}
To effectively capture the complementary information embedded in heterogeneous modalities, we design modality-aware processing branches tailored for the infrared and visible modalities.

The visible modality is typically characterized by rich high-frequency details, and convolutional operations, with their local inductive bias, are naturally suited for capturing these local patterns.
However, our PCA analysis reveals that the feature channels exhibit a certain degree of redundancy, suggesting that task-relevant signals are concentrated in specific subspaces.
Motivated by this observation, we propose the Split-Fuse Enhancer to efficiently amplify local discriminative signals in the visible flow.
As shown in Fig. \ref{fig:ho}, the visible representation $\bm{\mathcal{M}}_{vis}$ is split along the channel dimension into two groups with a ratio of $1/r$.
The selected group undergoes a $3 \times 3$ depth-wise convolution to extract local contextual information, while the remaining group retains the original features to preserve information integrity.
These two parts are then concatenated via a residual connection.
Finally, two $1 \times 1$ point-wise convolutions are applied to aggregate cross-channel information and project the features back to the unified latent space, yielding the enhanced representation $\bm{\mathcal{M}}^{e}_{vis}\in\mathbb{R}^{N \times \frac{C}{\gamma}}$.

In contrast, the infrared modality is dominated by low-frequency thermal radiation distributions, which provide coarse-grained object structures rather than fine-grained textures.
Applying local convolutions indiscriminately to infrared features risks distorting these global thermal structures or introducing high-frequency noise, which effectively undermines the unique physical priors of infrared data.
Therefore, we adopt a conservative strategy for the infrared branch, where the infrared representation $\bm{\mathcal{M}}_{\mathcal{P}}$ remains unaltered after the linear projection.
This design explicitly prevents the texture bias of convolutions from corrupting the low-frequency thermal cues, ensuring that the subsequent fusion stage receives clean and complementary multi-modal signals.

\begin{table*}[t]
\centering
\small
\caption{Overall performance on the VT821 \cite{tang2019rgbt}, VT1000 \cite{tu2019rgb} and VT5000 \cite{tu2022rgbt} datasets for IR-VIS salient object detection. The trainable parameters in backbones (\#TP), S-measure ($S_{\alpha}$), weighted F-measure ($F^{\omega}_{\beta}$), E-measure ($E_{m}$) and Mean Absolute Error ($MAE$) are reported. We use \textit{$\times$2} denotes a dual-branch backbone model.* denotes the results extended by us.}
\resizebox{1.0\textwidth}{!}
{
\renewcommand\arraystretch{1.1}
\begin{tabular}{l|c|c|c|cccc|cccc|cccc}
\toprule
\multirow{2}{*}{Methods}
& \multirow{2}{*}{Publications}
& \multirow{2}{*}{Backbone}
& \multirow{2}{*}{\#TP (M)}
& \multicolumn{4}{c|}{{VT821 \cite{tang2019rgbt}}}
& \multicolumn{4}{c|}{VT1000 \cite{tu2019rgb}} 
& \multicolumn{4}{c}{VT5000 \cite{tu2022rgbt}} \\ 
\cline{5-16}
&
&
&
& $S_{\alpha}\uparrow$ & $F^{\omega}_{\beta}\uparrow$ & $E_{m}\uparrow$ & $MAE\downarrow$
& $S_{\alpha}\uparrow$ & $F^{\omega}_{\beta}\uparrow$ & $E_{m}\uparrow$ & $MAE\downarrow$
& $S_{\alpha}\uparrow$ & $F^{\omega}_{\beta}\uparrow$ & $E_{m}\uparrow$ & $MAE\downarrow$ \\
\midrule
SwinNet \cite{liu2021swinnet}
& TCSVT 22
& Swin-B
& 88.0 \textit{$\times$2}
& 0.904 & 0.847 & 0.926 & 0.030 
& 0.938 & 0.896 & 0.947 & 0.018 
& 0.912 & 0.865 & 0.942 & 0.026 \\
LSNet \cite{zhou2023lsnet}
& TIP 23
& MobileNet-v2 
& 3.4 \textit{$\times$2}
& 0.877 & - & 0.911 & 0.033 
& 0.924 & - & 0.936 & 0.022 
& 0.876 & - & 0.916 & 0.036 \\
% CAVER \cite{pang2023caver}
% & TIP 23
% & ResNet-101
% & 44.5 \textit{$\times$2}
% & 0.898 & 0.845 & - & 0.027 
% & 0.938 & 0.911 & - & 0.017 
% & 0.899 & 0.849 & - & 0.028 \\
CCAFusion \cite{wang2024alignment}
& TSCVT 24 
& ResNet-50
& 23.6 \textit{$\times$2}
& 0.845 & 0.765 & - & 0.057 
& 0.911 & 0.872 & - & 0.027 
& 0.847 & 0.782 & - & 0.048 \\
% SACNet \cite{wang2024alignment}
% & TMM 24 
% & Swin-B
% & 88.0 \textit{$\times$2}
% & 0.906 & 0.859 & - & 0.025 
% & 0.942 & 0.927 & - & 0.014 
% & 0.917 & 0.888 & - & 0.021 \\
TCINet \cite{lv2024transformer}
& TCE 24
& Swin-B 
& 88.0
& 0.914 & 0.872 & 0.938 & 0.024 
& 0.942 & 0.920 & 0.968 & 0.015 
& 0.918 & 0.879 & 0.949 & 0.023 \\
TCINet-Swin-L* \cite{lv2024transformer}
& TCE 24
& Swin-L 
& 192.5
& 0.912 & 0.879 & 0.945 & 0.024 
& 0.942 & \textbf{0.931} & \textbf{0.975} & 0.013 
& 0.924 & 0.903 & 0.962 & \textbf{0.019} \\
TCINet-EVA02-L* \cite{lv2024transformer}
& TCE 24
& EVA02-L 
& 304.2
% sm, wfm, adaem, mae
& 0.843 & 0.766 & 0.895 & 0.042
& 0.884 & 0.839 & 0.928 & 0.034
& 0.823 & 0.742 & 0.887 & 0.050 \\
UniRGB-IR \cite{yuan2024unirgb}
& ACMM 25
& ViT-B 
& 8.9
& 0.881 & - & - & 0.039 
& 0.939 & - & - & 0.018 
& 0.906 & - & - & 0.027 \\
ConTriNet \cite{tang2024divide}
& TPAMI 25
& ResNet-50 
& 88.0 
& 0.878 & 0.817 & 0.918 & 0.033
& 0.925 & 0.899 & 0.960 & 0.020 
& 0.887 & 0.837 & 0.931 & 0.032 \\
ConTriNet-Swin-L* \cite{tang2024divide}
& TPAMI 25
& Swin-L 
& 192.5
& 0.862 & 0.788 & 0.906 & 0.037
& 0.911 & 0.871 & 0.941 & 0.025
& 0.861 & 0.789 & 0.909 & 0.040 \\
ConTriNet-EVA02-L* \cite{tang2024divide}
& TPAMI 25
& EVA02-L 
& 304.2 
& 0.852 & 0.776 & 0.899 & 0.040 
& 0.890 & 0.841 & 0.928 & 0.032 
& 0.839 & 0.758 & 0.895 & 0.047 \\
\midrule
CPNet (baseline) 
& IJCV 24
& Swin-L
& 192.5
& 0.890 & 0.831 & 0.917 & 0.033 
& 0.939 & 0.918 & 0.966 & 0.015 
& 0.911 & 0.873 & 0.949 & 0.025 \\
\rowcolor{gray!20}\textbf{+IV-tuning (Ours)}
& -
& Swin-L 
& \textbf{5.0}
& 0.904
& 0.848
& 0.928
& 0.029

& 0.940
& 0.915
& 0.964 
& 0.016

& 0.917
& 0.884
& 0.957
& 0.022\\
\midrule
CPNet (baseline)
& IJCV 24
& EVA02-L
& 304.2
& 0.890 & 0.832 & 0.915 & 0.034 
& 0.938 & 0.915 & 0.959 & 0.017 
& 0.907 & 0.866 & 0.943 & 0.026 \\
\rowcolor{gray!20}\textbf{+IV-tuning (Ours)}
& -
& EVA02-L 
& 7.6
& \textbf{0.924}
& \textbf{0.890}
& \textbf{0.951}
& \textbf{0.021}

& \textbf{0.948}
& \textbf{0.931}
& 0.971
& \textbf{0.012}

& \textbf{0.931}
& \textbf{0.904}
& \textbf{0.964}
& \textbf{0.019} \\
\bottomrule
\end{tabular}
}
\label{tab:sod}
\end{table*}

\begin{table*}[htbp]
    \centering
    \scriptsize  % 统一字号
    % 第一个表格：语义分割（用minipage替代subtable）
    \vspace{-10pt}
    \begin{minipage}[t]{0.48\textwidth}
        \centering
        \setlength{\tabcolsep}{1.2mm}    % 列间距
        \renewcommand\arraystretch{0.9}  % 行高
        \caption{Overall performance on the MFNet \cite{ha2017mfnet} dataset for IR-VIS semantic segmentation.}
        \label{tab:seg}  % 子表格单独label（可选）
        \hspace{-30pt}
        \begin{tabular}{l|c|c|c|c}  % 5列，匹配表头
        \toprule
        Methods & Publications & Backbone & \#TP (M) & \textbf{$\emph{m}$IoU} \\
        \midrule
        % RSFNet \cite{li2024residual} & ICCV 23 & ResNet-18  & 11.7 $\times$2 & 54.60  \\
        LASNet \cite{li2022rgb} & TCSVT 23 & ResNet-152  & 60.49 $\times$2 & 54.90  \\
        CDDFuse \cite{zhao2023cddfuse} & CVPR 23 & DeepLab-V3  & 42.5 $\times$2 & 44.50\\
        CMX \cite{zhang2023cmx}  & T-ITS 23 & MiT-B4 & 64.0 $\times$2 & 59.70 \\
        CAINet \cite{lv2024context} & TMM 24 & MobileNet-V2   & 3.4 $\times$2 & 58.60 \\
        CMX-Swin-L* \cite{zhang2023cmx}  & T-ITS 23 & Swin-L & 192.5 $\times$2 & 59.02 \\
        CAINet-Swin-L* \cite{lv2024context} & TMM 24 & Swin-L  & 192.5 $\times$2 & 60.09 \\
        CAINet-EVA02-L* \cite{lv2024context} & TMM 24 & EVA02-L  & 304.2 $\times$2 & 57.47 \\
        CMX-EVA02-L* \cite{zhang2023cmx}  & T-ITS 23 & EVA02-L & 304.2 $\times$2 & 59.79 \\
        % UniRGB-IR \cite{yuan2024unirgb} & ACMM 25 & ViT-B  & 8.9 & 59.30 \\
        \midrule
        Segformer (baseline) & NeurIPS 21  & EVA02-L & 304.2 & 54.53\\
        \rowcolor{gray!20}\textbf{+IV-tuning (Ours)} & - & EVA02-L & 7.6 & 60.14\\
        \midrule
        Segformer (baseline)  & NeurIPS 21 & Swin-L  & 192.5 & 56.78  \\
        \rowcolor{gray!20}\textbf{+IV-tuning (Ours)} & - & Swin-L  & \textbf{5.0} & \textbf{60.44} \\
        \bottomrule
        \end{tabular}
    \end{minipage}
    % \hfill  % 两个minipage之间添加自动间距（避免重叠）
    % 第二个表格：目标检测
    \begin{minipage}[t]{0.48\textwidth}
        \centering
        \setlength{\tabcolsep}{0.42mm}  % 列数多，缩小列间距
        \renewcommand\arraystretch{0.95}
        \caption{Overall performance on the M3FD \cite{liu2022target} dataset for IR-VIS object detection.}
        \label{tab:det}  % 子表格单独label（可选）
        \begin{tabular}{l|c|c|c|c|c|c}  % 7列，匹配表头
        \toprule
        Methods & Publications & Backbone & \#TP (M) & \textbf{$\emph{m}$AP} & \textbf{$\emph{m}$AP@50} & \textbf{$\emph{m}$AP@75} \\
        \midrule
        TarDAL \cite{liu2022target}  & CVPR 22 & CSPDarknet & 86.3 $\times$2  & 54.5  & 80.9  & - \\
        CDDFuse \cite{zhao2023cddfuse}  & CVPR 23 & CSPDarknet  & 86.3 $\times$2  & 54.6  & 81.1  & 57.0 \\
        % CBAM \cite{deevi2024rgb} & WACV 24 & EfficientNet  & 77.0 $\times$2  & 50.5  & 81.0  & 54.9 \\
        DAMSDet \cite{guo2024damsdet} & ECCV 24 & ResNet-50  & 23.6 $\times$2  & 47.8  & 75.8  & 49.6 \\
        CoCoNet \cite{liu2023coconet} & IJCV 24 & CSPDarknet  & 86.3 $\times$2  & 54.2  & 80.7  & - \\
        CoCoNet-Swin-L* \cite{liu2023coconet} & IJCV 24 & Swin-L  & 192.5 $\times$2  & 57.5  & 83.1  & 58.6 \\
        ICAFusion \cite{shen2024icafusion}  & PR 24  & CSPDarknet  & 86.3 $\times$2  & 55.1  & 84.0  & 53.2 \\
        ICAFusion-Swin-L* \cite{shen2024icafusion}  & PR 24  & Swin-L  & 192.5 $\times$2  & 60.3  & 89.2  & 65.4 \\
        WaveFusion \cite{wang2025wavefusion} & TCSVT 25 & CSPDarknet & 86.3 $\times$2  & -  & 87.5  & - \\
        \midrule
        DINO (baseline)  & ICLR 23 & Swin-L & 192.5  & 60.1  & 90.8  & 64.0 \\
        \rowcolor{gray!20}\textbf{+IV-tuning (Ours)} & - & Swin-L  & \textbf{5.0}  & 61.2  & 90.9  & \textbf{66.6} \\
        CO-DETR (baseline) & ICCV 23 & Swin-L & 192.5  & 60.5  & 91.0  & 64.1 \\
        \rowcolor{gray!20}\textbf{+IV-tuning (Ours)} & - & Swin-L  & \textbf{5.0}  & \textbf{62.1}  & \textbf{91.7}  & 65.9 \\
        \bottomrule
        \end{tabular}
    \end{minipage}
\end{table*}

\textit{3) Rank-aware Fusion:}
As corroborated by our PCA analysis (Fig. \ref{fig:pca} (b)), the feature space exhibits a distinct phase transition after entering the frozen PVM, where the explained variance ratio of the dominant component drops sharply (0.293 $\rightarrow$ 0.126).
This observation aligns with the theory of Intrinsic Dimensionality \cite{ansuini2019intrinsic}, suggesting that shallow layers encode redundant, low-level patterns with a lower intrinsic dimension, whereas deep layers evolve into semantic-rich manifolds.

Driven by this insight, we tailor both the hidden dimensions and fusion strategies to match these varying intrinsic properties.
Specifically, for MP-$\alpha$, we employ a highly compressed subspace ($\frac{C}{\gamma_1}=8$) and perform fusion directly within this latent space.
Given the low intrinsic dimension, this strategy efficiently forces the alignment of coarse-grained modalities without information loss.
For MP-$\beta$, we increase the hidden dimension ($\frac{C}{\gamma_2}=64$) and project features back to the high-dimensional space $C$ before fusion.
This simple design explicitly preserves the structural independence of each modality's complex semantic manifold, thereby avoiding information distortion during cross-modal interaction.
In summary, the forward process of $\alpha$ and $\beta$ fusion can be formulated as:
\begin{align}
    \bm{\mathcal{P}}^{0} = s_{3}(\bm{\mathcal{M}}^{e}_{vis} + \bm{\mathcal{M}}_{\mathcal{P}}),\\
    \bm{\mathcal{P}}^{l} = s_{3}(\bm{\mathcal{M}}^{e}_{vis}) +s_{4}(\bm{\mathcal{M}}_{\mathcal{P}}),
\end{align}
where $s_{3}(\cdot)$ and $s_{4}(\cdot)$ denote the up-projection linear layers with the output dimension of $C$.

Finally, the generated prompt is merged with the backbone tokens through element-wise addition.
The initial prompt $\mathcal{P}^{0}$ is input into the next MP-$\beta$ block, while the output of each encoder layer is updated to $\mathcal{P}^{l}$, serving as the input for the next MP-$\beta$ block.
The two MP-$\beta$ blocks share weights within each encoder layer while remaining independent across layers, which we find reduces parameters with no performance loss.

\subsection{Optimization}
During the tuning process, we freeze the parameters of PVMs and only optimize a few parameters $\theta_{IVT}=\{\tau^{ir}, \mathcal{P}_{\alpha}, \{\mathcal{P}^{l}_{\beta}\}^{L}_{l=1}, \phi\}$, where $\tau^{ir}$ denotes the patch embedding layer of infrared inputs.
Therefore, the optimization process can be formulated as:
\begin{align}
\theta_{IVT}^* = \operatorname*{arg\,min}_{\theta_{IVT}} \frac{1}{|\mathcal{D}|} \sum_{(\bm{x}, \bm{y}) \in \mathcal{D}} \mathcal{L} \left( \phi \big( f(\bm{x}; \theta_{IVT}) \big), \bm{y} \right),
\end{align}
where $\bm{x}=\{\bm{x}_{vis}, \bm{x}_{ir}\}$ denotes the input pair, and $\mathcal{L}$ is the task-specific loss function consistent with the full fine-tuning baseline.
For semantic segmentation, we use cross-entropy loss. For salient object detection, we follow CPNet \cite{hu2024cross} and adopt a hybrid loss combining binary cross-entropy and IoU loss. 
For object detection, the loss function can be formulated as:
\begin{align}
    \mathcal{L}=L_{cls}+\lambda_{iou}L_{iou}+\lambda_{L_{1}}L_{1},
\end{align}
where $L_{cls}$ is the focal loss for classification, $L_{1}$ and generalized IoU loss $L_{iou}$ are employed for bounding box regression, with $\lambda_{iou}$ and $\lambda_{L_{1}}$ as the regularization parameters.

\section{Experiment}
\label{sec:experiment}
\subsection{Experiment Settings}
\vspace{-3pt}
\textit{1) Downstream Tasks:}
To verify the generality of IV-tuning, we conduct experiments on three mainstream IR-VIS high-level tasks:
(1) For salient object detection, we evaluate our method on the VT821 \cite{tang2019rgbt}, VT1000 \cite{tu2019rgb} and VT5000 \cite{tu2022rgbt} datasets.
We follow previous works to use VT5000’s 2,500 pairs for training and the remaining for testing.
We use the S-measure ($S_{\alpha}$), weighted F-measure ($F^{\omega}_{\beta}$), E-measure ($E_{m}$) and mean absolute error ($MAE$).
(2) For semantic segmentation, we report the results on the widely used MFNet \cite{ha2017mfnet} dataset.
We use the mean Intersection over Union (\emph{m}IoU) and mean Accuracy (\emph{m}Acc.) for evaluation.
(3) For object detection, we provide the comparison results on the M3FD \cite{liu2022target} dataset, the \emph{m}AP, \emph{m}AP50, and \emph{m}AP75 are employed for evaluation.

\textit{2) Pre-trained Visual Models and Decoders:}
Benefiting from the plug-and-play nature of our IV-tuning, we can directly adopt state-of-the-art decoders from the visible domain for different tasks without designing complex custom decoders.
For salient object detection and semantic segmentation, we use CPNet \cite{hu2024cross} and Segformer \cite{xie2021segformer} as decoders, respectively, with vanilla Swin Transformer \cite{liu2021swin} and EVA02 \cite{fang2024eva} as the PVMs.
For object detection, we only use the widely adopted Swin Transformer \cite{liu2021swin} as the PVM, and select CO-DETR \cite{zong2023detrs} and DINO \cite{zhang2022dino} as the detectors.
To balance precision and efficiency, we employ the large version of these architectures.

\begin{table*}[t]
\centering
\scriptsize
\begin{minipage}[b]{0.49\linewidth}
    \centering
    \renewcommand\arraystretch{0.9}
    \setlength{\tabcolsep}{1.2mm}
    \caption{Comparisons with PETL methods on the MFNet \cite{ha2017mfnet} and M3FD \cite{liu2022target} datasets.}
    \label{petl}
    \begin{tabular}{l|c|c|c c|c c}
        \toprule
        \multirow{3}{*}{Methods} & \multirow{3}{*}{Pub.} & \multirow{3}{*}{\#TP(M)} & \multicolumn{2}{c|}{MFNet \cite{ha2017mfnet}} & \multicolumn{2}{c}{M3FD \cite{liu2022target}} \\
        \cline{4-7}
        & & & \multicolumn{2}{c|}{Swin-L+Segformer} & \multicolumn{2}{c}{Swin-L+CO-DETR} \\
        \cline{4-7}
        & & & \emph{m}IoU & \emph{m}Acc. & \emph{m}AP & \emph{m}AP@75 \\
        \midrule
        Full Fine-tune & - & 192.5 & 56.78 & 64.88 & 60.5 & 64.1 \\
        Freeze & - & 0.0 & 53.02 & 60.66 & 56.6 & 58.2 \\
        \midrule
        +VPT \cite{jia2022visual} & ECCV 22 & 0.2 & 56.79 & 64.55 & 61.1 & 65.3 \\
        +LoRA \cite{hu2022lora} & ICLR 22 & 1.0 & 54.37 & 62.89 & 60.9 & 64.1 \\
        +AdaptFormer \cite{chen2022adaptformer} & NeurIPS 22 & 2.3 & 49.79 & 54.78 & 61.0 & 64.5 \\
        +Bi-AdaptFormer \cite{jie2023revisiting} & ICCV 23 & 1.2 & 58.14 & 65.28 & 61.7 & 64.7 \\
        +Rein \cite{wei2024stronger} & CVPR 24 & 14.7 & 57.83 & 65.07 & 61.6 & 65.3 \\
        +Mona \cite{yin20255} & CVPR 25 & 7.0 & 59.11 & 65.57 & 61.4 & 65.1 \\
        \rowcolor{gray!20}\textbf{+IV-tuning (Ours)} & - & 5.0 & \textbf{60.44} & \textbf{68.77} & \textbf{62.1} & \textbf{65.9} \\
        \bottomrule
    \end{tabular}
\end{minipage}
\hfill % 撑开左右间距
% 右边表格 (Table II)
\begin{minipage}[b]{0.49\linewidth}
    \centering
    \renewcommand\arraystretch{0.9}
    \setlength{\tabcolsep}{1.8mm} % 稍微调整列间距以适应宽度
    \caption{Comparisons with the Combination of IR-VIS Methods and PETL Methods on the MFNet Dataset.}
    \label{seg-petl+cmx}
    \begin{tabular}{l|c|c c}
        \toprule
        Methods & \#TP (M) & \emph{m}IoU & \emph{m}Acc. \\
        \midrule
        \multicolumn{4}{c}{Task: Semantic Segmentation / Backbone: Swin-L \cite{liu2021swin}} \\
        CAINet \cite{lv2024context} & 192.5 $\times2$ & 58.60 & 65.44  \\
        CAINet \cite{lv2024context} +LoRA & 1.0 $\times2$ & 52.85 & 60.97 \\
        CAINet \cite{lv2024context} +Bi-Adaptformer & 1.2 $\times2$ & 56.99 & 65.84 \\
        CAINet \cite{lv2024context} +Rein & 14.7 $\times2$ & 53.18 & 61.14  \\
        CAINet \cite{lv2024context} +Mona & 7.0 $\times2$ & 57.64 & 65.78  \\
        CMX \cite{qingyun2021cross} & 192.5 $\times2$ & 59.02 & 67.26  \\
        CMX \cite{qingyun2021cross} +LoRA & 1.0 $\times2$ & 50.44 & 57.21 \\
        CMX \cite{qingyun2021cross} +Bi-Adaptformer & 1.2 $\times2$ & 55.03 & 62.64 \\
        CMX \cite{qingyun2021cross} +Rein & 14.7 $\times2$ & 51.02 & 58.07  \\
        CMX \cite{qingyun2021cross} +Mona & 7.0 $\times2$ & 59.89 & 68.17  \\
        \rowcolor{gray!20}\textbf{IV-tuning (Ours)} & 5.0 & \textbf{60.44} & \textbf{68.77}\\
        \bottomrule
    \end{tabular}
\end{minipage}

\vspace{3pt} 
\begin{minipage}{1.0\linewidth}
    \centering
    \renewcommand\arraystretch{1.0}  % 恢复行高，保持清晰
    \setlength{\tabcolsep}{1.8mm}      % 适当放宽列间距
    \caption{Comparisons of trainable backbone parameters (\#TP), storage, GPU memory, speed and performance across multiple tasks.}
    \label{speed}
    \begin{tabular}{c c c l| c c c c c c} 
        \toprule
        Task & Method & DataType & Backbone-Head & \#TP & Storage & GPU Memory & Train Speed & Infer Speed & Performance\\
        \midrule
        Semantic Segmentation
        & Full Fine-tuning & VIS    & EVA02-L+Segformer & 304.2M            & 3.7 GB & 9.8 GB   & 27.0 h  & 6.4 fps & \emph{m}IoU: 56.78 \\
        Semantic Segmentation
        & Full Fine-tuning & VIS+IR & EVA02-L+Segformer & 304.2M $\times2$ & 6.8 GB  & 18.2 GB & 50.5 h & 3.6 fps    & \emph{m}IoU: 58.29 \\
        \rowcolor{gray!20} Semantic Segmentation & IV-tuning (Ours)        & VIS+IR & EVA02-L+Segformer & 7.5M              & 1.4 GB & 10.0 GB  & 33.0 h  & 4.8 fps  & \emph{m}IoU: 60.44 \\
        \midrule
        Object Detection
        & Full Fine-tuning & VIS    & Swin-L+CO-DETR    & 192.50M           & 2.7 GB  & 16.7 GB  & 28.1 h & 2.4 fps    & \emph{m}AP: 60.5 \\
        Object Detection
        & Full Fine-tuning & VIS+IR & Swin-L+CO-DETR    & 192.50M $\times2$ & 4.8 GB & 25.6 GB  & 55.1 h & 2.1 fps    & \emph{m}AP: 61.9 \\
        \rowcolor{gray!20} Object Detection & IV-tuning (Ours)        & VIS+IR & Swin-L+CO-DETR    & 5.0M              & 1.3 GB  & 19.2 GB  & 29.2 h & 2.6 fps    & \emph{m}AP: 62.1 \\
        \midrule
        Salient Object Detection
        & Full Fine-tuning   & VIS    & EVA02-L+CPNet & 304.2M            & 3.7 GB  & 20.6 GB  & 15.7 h & 13.1 fps & $S_\alpha$: 0.832 \\
        Salient Object Detection
        & Full Fine-tuning   & VIS+IR & EVA02-L+CPNet & 304.2M $\times2$ & 6.8 GB  & 33.3 GB  & 32.6 h & 6.4 fps & $S_\alpha$: 0.855 \\
        \rowcolor{gray!20} Salient Object Detection & IV-tuning (Ours)          & VIS+IR & EVA02-L+CPNet & 7.5M              & 1.4 GB  & 21.0 GB  & 15.8 h & 8.5 fps  & $S_\alpha$: 0.890 \\
        \bottomrule
    \end{tabular}
\end{minipage}

\end{table*}

\textit{3) Implementation Details:}
For salient object detection, we use Adam (lr = 5e-5, weight decay = 1e-1) and train for 200 epochs with a batch size of 8 on $384\times384$ cropped images.
For semantic segmentation, we use SGD (lr = 1e-3, weight decay = 1e-2) and train for 160,000 iterations with a batch size of 2 on $512 \times 512$ cropped images.
For object detection, we use AdamW (lr = 1e-4, weight decay = 1e-4) and train for 48 epochs with a batch size of 2 on $640\times640$ cropped images.
The baseline method is constructed by fully fine-tuning the model using only the visible modality.
All experiments follow identical settings for fair comparison.
During tuning, the backbone is frozen, while the decoder and the IV-tuning components are trainable.
All experiments are conducted on the RTX 3090.
For fair comparison, we extend selected IR-VIS methods's backbones while following their original settings, and the extended results are denoted with * in Tables \ref{tab:sod}, \ref{tab:seg}, and \ref{tab:det}.
\subsection{Comparison With State-of-the-Art IR-VIS Methods}
\textit{1) IR-VIS Salient Object Detection}:
As shown in Table \ref{tab:sod}, without special designs, the Swin-L-equipped baseline with only visible data achieves competitive performance against LSNet \cite{zhou2023lsnet}.
This indicates that transferring the powerful PVMs is a promising approach.
With infrared modality incorporated, IV-tuning outperforms them with far fewer trainable backbone parameters.
Notably, most extended versions of TCINet \cite{lv2024transformer} and ConTriNet \cite{tang2024divide} suffer varying degrees of performance degradation.
This reveals that traditional IR-VIS methods are inherently incompatible with advanced large-scale PVMs, which we attribute to the intrinsic modality gap between infrared and visible modalities, as these methods adopt two weight-sharing backbones for dual-modal learning.
In contrast, IV-tuning leverages a single backbone to facilitate cross-modal complementarity learning, achieving superior generalization with fewer trainable parameters.

\textit{2) IR-VIS Semantic Segmentation:}
As shown in Table \ref{tab:seg}, IV-tuning surpasses the full fine-tuning baselines, gaining an improvement of 10.3\% and 6.4\% when coupled with EVA02-L and Swin-L, respectively.
% Besides, IV-tuning outperforms the state-of-the-art methods, achieving the highest \emph{m}IoU of 60.44$\%$.
Furthermore, the extended CMX \cite{zhang2023cmx} suffered from severe overfitting, leading to the loss of generalization ability (e.g., CMX (59.97) vs. CMX-Swin-L (59.02)).
In contrast, IV-tuning achieved the highest \emph{m}IoU of 60.44 with only 5.0M trainable backbone parameters, realizing dual benefits of generalization and efficiency.

\textit{3) IR-VIS Object Detection:}
As shown in Table \ref{tab:det}, benefiting from the supplementary information provided by the infrared modality, IV-tuning achieves improvements of 2.8\% and 4.1\% in \emph{m}AP\@75 on Swin-L+CO-DETR and Swin-L+DINO, respectively.
This indicates that IV-tuning effectively leverages the complementary information from visible and infrared modalities, resulting in more precise predictions of object shapes and boundaries under high IoU conditions.
Meanwhile, IV-tuning achieves a 2.0\% improvement in \emph{m}AP compared to ICAFusion \cite{shen2024icafusion}, demonstrating the advantages of IV-tuning in harnessing large-scale PVMs for object detection.

\subsection{Comparisons with State-of-the-Art Parameter-Efficient Transfer Learning Methods}
\textit{1) Adapting PETL methods for IR-VIS Tasks:}
We adapt representative PETL methods for IR-VIS tasks by introducing an independent infrared patch embedding layer and summing the tokens from both modalities.
Experiments in Table \ref{petl} demonstrate that our IV-tuning achieves superior performance with modest trainable parameters.
This validates that a tailored design is essential for maximizing the potential of PETL in multi-modal scenarios.

\begin{table}
    \centering
    \scriptsize
    \renewcommand\arraystretch{1.0} % 全局统一行高，避免局部重复设置
    \vspace{-5pt}
    % ========== 左侧表格：核心改基线+删硬调hspace+精准宽度 ==========
    \begin{minipage}[t]{0.5\textwidth}  % 调宽为0.45，与右侧0.5搭配更合理
        \centering
        \setlength{\tabcolsep}{2.7mm}
        % 核心修复：用\vtop将基线强制拉到minipage最顶部，与右侧严格对齐
        \vtop{
        \caption{Performance Results on More Pre-trained Visual Models.}
        \begin{tabular}{c c | c c c}
        \toprule
        Backbone-Head & Method & \#TP & \emph{m}IoU & \emph{m}Acc.\\
        \midrule
        CLIP-L+Segformer & Full Fine-tuning & 316.7 M & 49.40 & 56.54\\
        \rowcolor{gray!20}CLIP-L+Segformer & IV-tuning (Ours) & \textbf{7.5 M} & \textbf{50.23} & \textbf{57.32}\\
        \midrule
        MAE-L+Segformer & Full Fine-tuning & 306.5 M & 53.51 & 61.00 \\
        \rowcolor{gray!20}MAE-L+Segformer & IV-tuning (Ours) & \textbf{7.5 M} & \textbf{55.34} & \textbf{63.35} \\
        \midrule
        SAM-H+Segformer & Full Fine-tuning & 632.2 M & 58.71 & 67.54 \\
        \rowcolor{gray!20}SAM-H+Segformer & IV-tuning (Ours) & \textbf{12.2 M} & \textbf{59.65} & \textbf{68.53} \\
        \midrule 
        DINOv3-L+Segformer & Full Fine-tuning & 306.5 M & 59.06 & 68.45 \\
        \rowcolor{gray!20}Dinov3-L+Segformer & IV-tuning (Ours) & \textbf{7.5 M} & \textbf{60.14} & \textbf{68.79} \\
        \bottomrule
        \end{tabular}
        \label{tab:morepvms}
        }
    \end{minipage}
    % \quad  % 左右固定间距（0.5em左右，比\hfill可控，无偏移）
    % ========== 右侧表格：仅微调宽度，其余完全保留 ==========
    \begin{minipage}[t]{0.5\textwidth}
        \centering
        % 第一个子表：Variants of IV-tuning
        \setlength{\tabcolsep}{0.7mm}
        \caption{Variants of IV-tuning.}
        \begin{tabular}{lccccc}
        \toprule
        Variants & (a) Symmetric1 & (b) Symmetric2 & (c) VIS-only & (d) Uni-fusion & IV-tuning \\
        \midrule
        \#TP (M) & 4.82 & 5.23 & 5.23 & 3.46 & 5.03 \\
        \emph{m}IoU & 59.31 & 58.98 & 56.82 & 59.07 & \textbf{60.44} \\
        \bottomrule
        \end{tabular}
        \label{variants_tab}
        \vspace{0.8em}
        % 第二个子表：Ablation on split ratio
        \setlength{\tabcolsep}{3.7mm}
        \caption{Ablation studies on the split ratio.}
        \begin{tabular}{lccccc}
        \toprule
        Split ratio $\frac{1}{r}$ & 0.00 & 0.25 & 0.50 & 0.75 & 1.00 \\
        \midrule
        \#TP (M) & 5.023 & 5.026 & 5.030 & 5.034 & 5.038 \\
        \emph{m}IoU & 59.46 & \textbf{60.44} & 59.65 & 60.16 & 59.94 \\
        \bottomrule
        \end{tabular}
        \label{split}
    \end{minipage}
\end{table}

\textit{2) Integrating PETL into Existing IR-VIS Methods:}
An alternative strategy for IR-VIS tasks involves integrating PETL methods into existing IR-VIS frameworks to achieve efficient transfer learning.
Hence, we integrate advanced PETL modules into dual-branch IR-VIS frameworks, as shown in Table \ref{seg-petl+cmx}.
However, the results reveal that this naive combination may leads to significant degradation (e.g., CMX drops 7.99\% in \emph{m}IoU).
We attribute this failure to two factors:
First, the texture-biased design of standard PETL mismatches the low-frequency nature of infrared data, which is better preserved by linear projections (as analyzed in Section \ref{sec:intro}).
Second, independent tuning of dual branches disrupts inter-modal feature alignment.
Restricting updates to separate modules fails to synchronize feature spaces across branches, leading to misalignment that impedes effective fusion.
In contrast, IV-tuning adopts a unified single-branch architecture with a Modality-aware Prompter, effectively ensuring cross-modal synchronization and superior fusion.

\vspace{-5pt}
\subsection{Comparisons of Computational Efficiency}
Table \ref{speed} details the efficiency comparison against the full fine-tuning paradigm.
While IV-tuning incurs a slight overhead compared to single-modal baselines due to the additional infrared input, it significantly outperforms the dual-branch full fine-tuning baseline in efficiency.
Specifically, IV-tuning reduces training GPU memory usage by 45.1\%, 25.0\%, and 36.9\% across the three tasks, respectively.
This efficiency also minimizes storage costs, as only one copy of the backbone parameters plus $n$ sets of lightweight task-specific parameters is required for $n$ downstream tasks.
Furthermore, IV-tuning achieves faster inference speeds and alleviates the overfitting inherent in dual-branch structures, leading to better generalization.

\begin{figure}[t]
  \centering
  \includegraphics[width=1.0\linewidth]{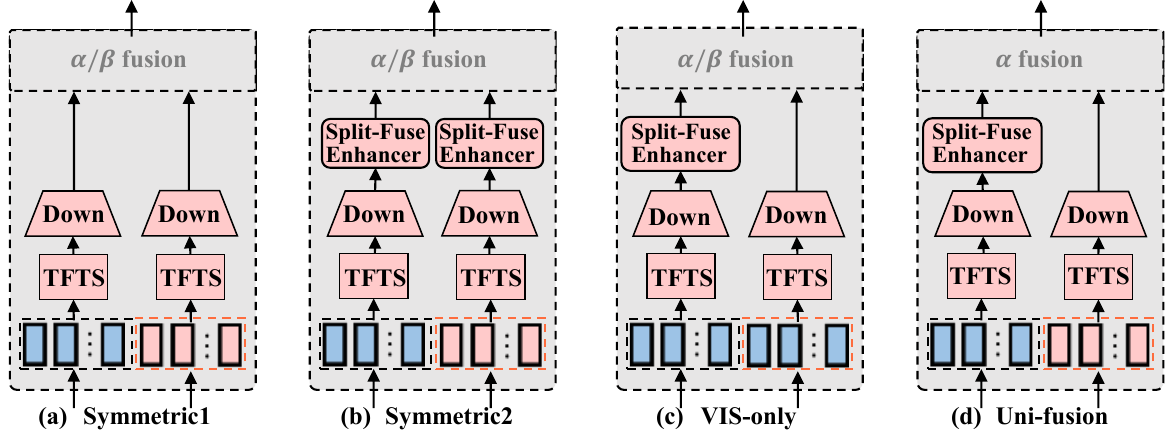}
  \caption{Variants of MP blocks. Corresponding to Table \ref{variants_tab}, these results provide empirical support for the design of our modality-aware process and the effectiveness of the proposed $\alpha$ / $\beta$ fusion strategy.}
  \label{variants}
\end{figure}

\begin{figure}[t]
  \centering
  \vspace{-5pt}
  \includegraphics[width=1.0\linewidth]{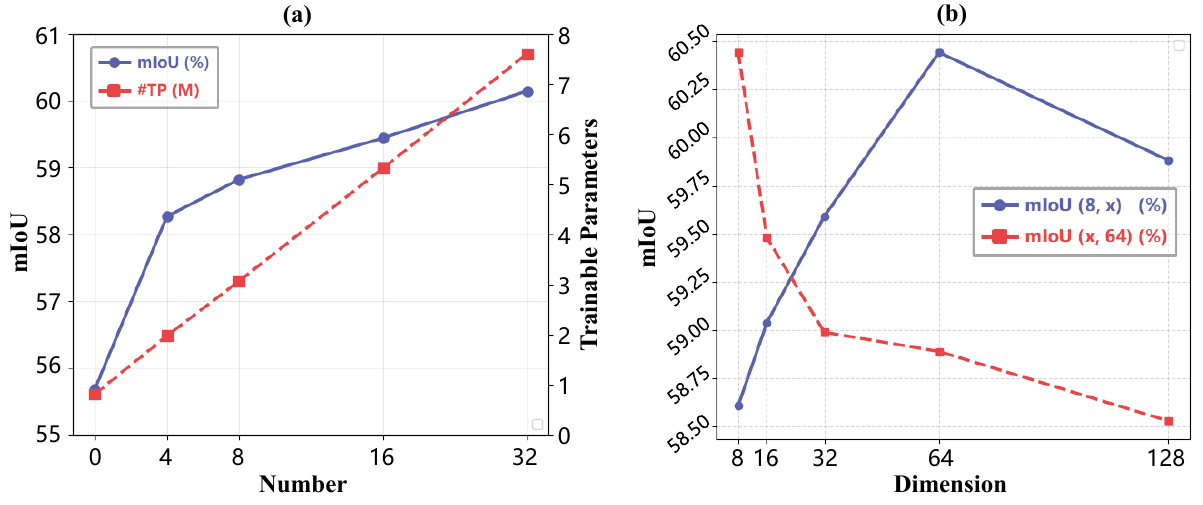}
  \caption{Influence on the number of MP-$\beta$ blocks (a) and dimension combinations (b). In (a), a zero count of MP-$\beta$ corresponds to the ablation study that uses MP-$\alpha$ alone.
  }
  \vspace{-5pt}
  \label{dim-num}
\end{figure}

\subsection{Ablation Study}
\textit{1) Variants Comparisons:}
To validate the rationale behind our module design, we conduct variant studies of the MP block on the MFNet dataset with Swin-L+Segformer \cite{liu2021swin,xie2021segformer}.
As shown in Table \ref{variants_tab} and Fig. \ref{variants}, 
(a and b) investigate the impact of the Feature Enhancer on each flow.
We observe that adding convolution modules to the infrared flow impairs performance, while slightly improving it in the visible flow.
This supports our choice to preserve the infrared's low-frequency characteristics via linear projection.
(c): when the infrared is missing, our method degrades to visible-based PETL, yet still outperforms the full Fine-tuning baseline.
This demonstrates that the performance gain of IV-tuning partly stems from the suppression of overfitting and partly from the integration of complementary information between infrared and visible modalities.
(d): performance degradation with a unified fusion method (59.07 vs. 60.44) highlights the effectiveness of our tailored fusion strategy for different feature spaces, where a greater flexibility is essential for high-rank feature spaces to preserve modality-specific characteristics.

\begin{figure*}[t]
  \centering
  \includegraphics[width=1.0\linewidth]{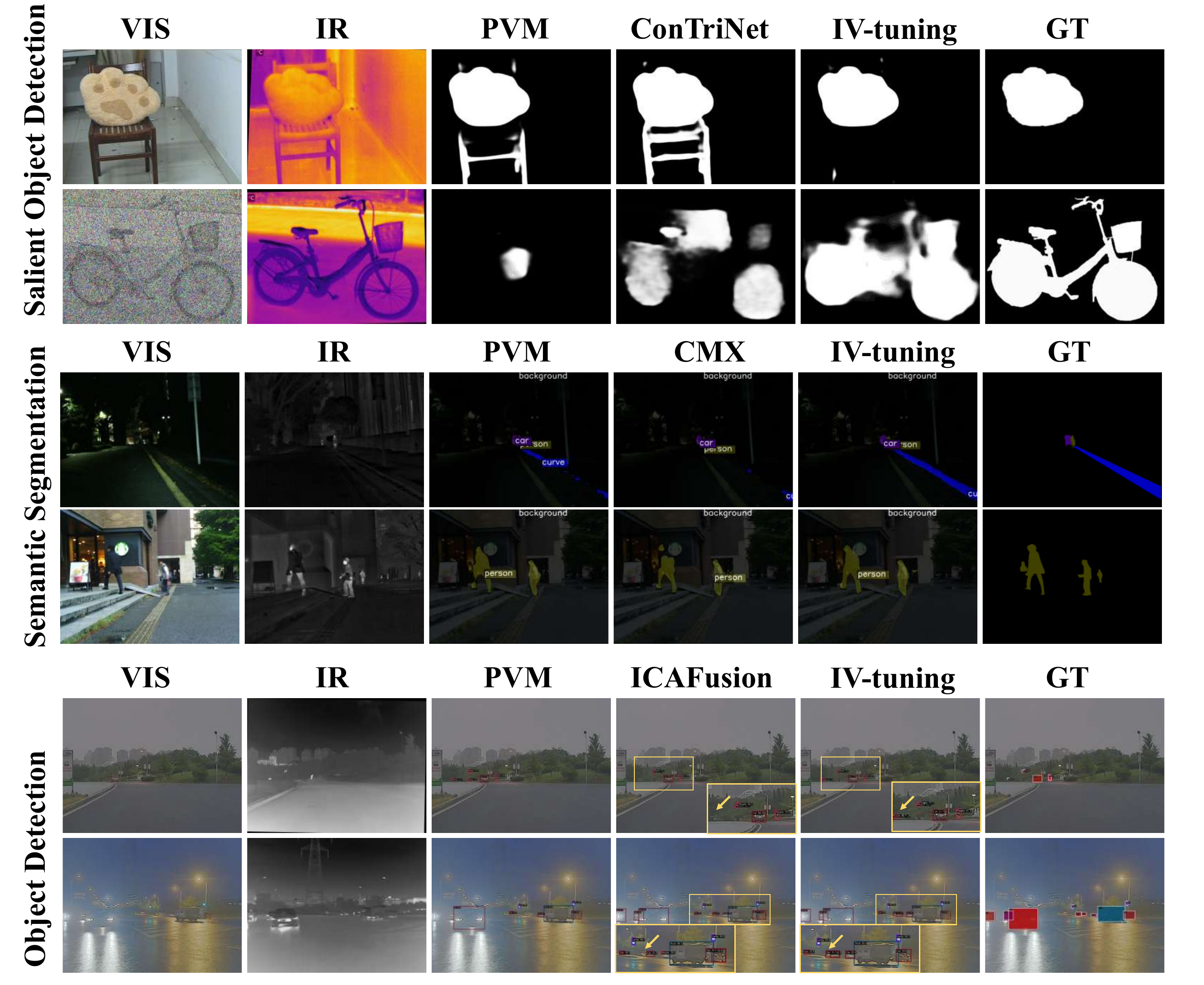}
  \caption{Visual comparisons of results across three tasks, including salient object detection, semantic segmentation and object Detection.
}
  \label{Visualizations3task}
\end{figure*}

\begin{figure}[t]
  \centering
  \includegraphics[width=1.0\linewidth]{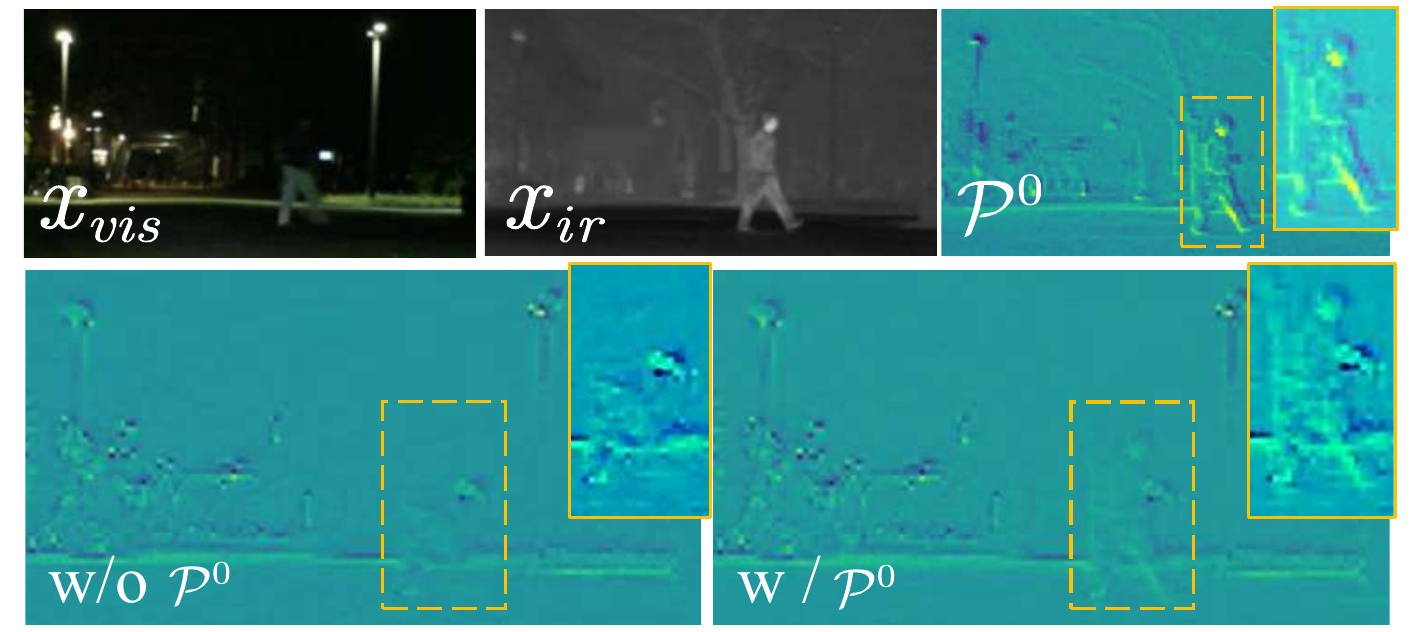}
  \caption{Visualization of the prompting process. To enable easier visual inspection and comparison, we apply equal enhancement to both contrast and brightness on the target regions.
}
  \label{visual}
\end{figure}

\textit{2) Ablation Study on the Block Number and Dimensional Combinations:}
We investigate the impact of the number of MP-$\beta$ and dimensional combinations on the MFNet \cite{ha2017mfnet} dataset with Swin-L+Segformer \cite{liu2021swin,xie2021segformer}.
As shown in Fig. \ref{dim-num} (a), increasing the number of inserted MP-$\beta$ blocks consistently improves performance, so we adopt full-layer insertion by default.
When the number of MP-$\beta$ blocks is set to zero, our method utilizes only the MP-$\alpha$, analogous to the ``shallow" setting in VPT \cite{jia2022visual}.
This configuration achieves an mIoU of 55.63\%, closely approaching the full fine-tuning baseline of 56.78\%.
This demonstrates that a single MP-$\alpha$ module, despite introducing fewer than 1M trainable parameters, is sufficient to yield performance comparable to full fine-tuning.
We further analyze the impact of different hidden dimensions for MP-$\alpha$ and MP-$\beta$ (Fig. \ref{dim-num} (b)). Interestingly, larger dimensions benefit MP-$\beta$ but harm MP-$\alpha$, indicating that a smaller dimension can effectively capture concentrated principal components in the low-ranked feature space, while diverse spaces require more flexibility to avoid losing effective information.

\textit{3) Ablation Study on the Split Ratio of the Split-Fuse Enhancer:}
We report the effect of different split ratios in Table \ref{split}.
Varying the ratio leads to slight parameter changes and minimal performance fluctuation.
Therefore, we adopt a 1/4 split ratio by default.

\begin{table}[htbp]
    \centering
    \scriptsize
    \setlength{\tabcolsep}{0.7mm} % 1.2mm
    \caption{Comparisons on the NYUDepthv2 Dataset \cite{Silberman:ECCV12}.}
    \renewcommand\arraystretch{1.0}
    \begin{tabular}{c c c| c c c}
    \toprule
    Backbone-Head & Method & Publications & \#TP (M) & \emph{m}IoU & \emph{m}Acc.\\
    \midrule
    MiT-B4+CMNext & CMNext \cite{zhang2023delivering} & CVPR 23 & 119.6 & 56.9 & - \\
    MiT-B5+GeminiFusion & GeminiFusion \cite{jia2024geminifusion} & ICML 24 & 137.2 & 57.7 & - \\
    DFormerv2-L+Hamburger & Dformerv2 \cite{dformerv2} & CVPR 25 & 95.5 & 58.40 & -\\
    \midrule
    Swin-L+Segformer & Full Fine-tuning & - & 192.5 & 54.75 & 69.00 \\
    \rowcolor{gray!20}Swin-L+Segformer  & IV-tuning (Ours) & -& \textbf{5.0} & \textbf{56.34} & \textbf{69.18} \\
    \midrule
    EVA02-L+Segformer & Full Fine-tuning & - &  304.2 & 63.28 & 75.47\\
    \rowcolor{gray!20}EVA02-L+Segformer & IV-tuning (Ours) & -  & \textbf{7.5} & \textbf{66.52} & \textbf{78.03}\\
    \bottomrule
    \end{tabular}
    \label{tab:rgbd}
\end{table}

\subsection{Qualitative Results}
We first present the visual comparison of results on the three tasks in Fig. \ref{Visualizations3task}.
In the salient object detection task, it can be clearly observed that the pre-trained vision model with full fine-tuning and ConTriNet \cite{tang2024divide} exhibit overfitting to object characteristics, e.g., the ``chair" in the first row.
In contrast, IV-tuning demonstrates favorable generalization ability.
Furthermore, benefiting from the ability to learn the complementarity of both modalities, IV-tuning achieves better performance in the semantic segmentation task, yielding more accurate segmentation boundaries. 
Meanwhile, in the object detection task, it effectively reduces false alarms and missed detections for small objects.

Then, we visualize to illustrate how the modal prompt learns and injects complementary information.
Modal prompt $\mathcal{P}^{0}$ effectively captures complementary information from both infrared and visible modalities, as exemplified by the pedestrian region in Fig. \ref{visual}. By infusing the backbone with fine-grained, modality-aware details, it yields more discriminative feature representations and ultimately leads to enhanced task performance.

\begin{figure}[ht]
  \centering
  \includegraphics[width=1\linewidth]{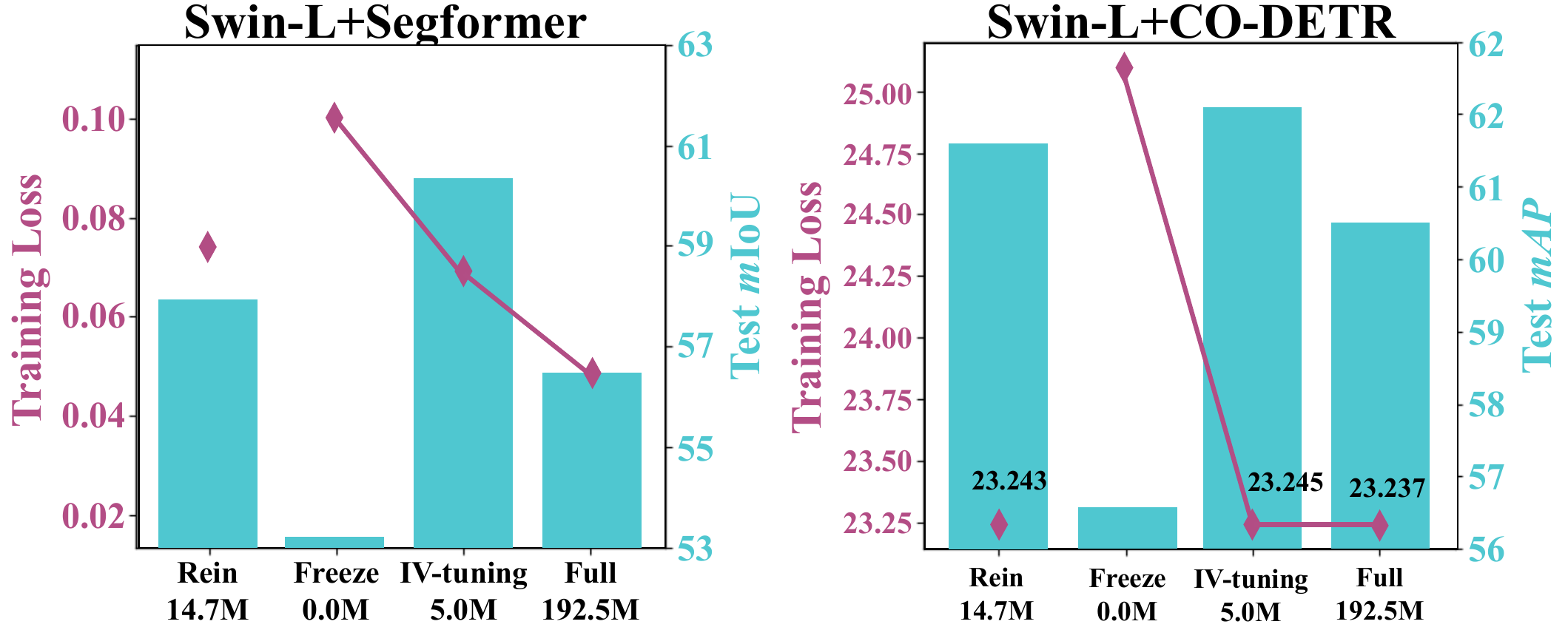}
  \caption{Training loss and test metrics on the MFNet \cite{ha2017mfnet} and M3FD \cite{liu2022target} datasets, respectively.
  We use Swin-L+Segformer \cite{liu2021swin,xie2021segformer} for semantic segmentation and Swin-L+CO-DETR \cite{liu2021swin,zong2023detrs} for object detection, respectively.
  The purple line denotes the average training loss when fine-tuned on the MFNet \cite{ha2017mfnet} and M3FD \cite{liu2022target} datasets, respectively.
  The blue bars are the test metrics, including \emph{m}IoU and \emph{m}AP, respectively.}
  \label{fig:loss and metric}
\end{figure}

\subsection{Scalability Analysis}
\textit{1) More Pretrained Visual Models:}
We further validate the effectiveness of our proposed method on additional Pretrained Visual Models (PVMs), including CLIP \cite{radford2021learning}, MAE \cite{he2022masked}, SAM \cite{kirillov2023segment} and DINOv3 \cite{simeoni2025dinov3}, as shown in Fig.~\ref{tab:morepvms}.
It is noteworthy that even for PVM that performs suboptimally in the IR-VIS domain (e.g., CLIP), our IV-tuning approach surpasses full fine-tuning while requiring fewer parameters.
It is also worth emphasizing that the upper bound of IV-tuning is inherently tied to the capability of the underlying PVM theoretically.
As more powerful PVMs continue to emerge, IV-tuning provides a promising and efficient pathway to transfer their pre-trained knowledge into the IR-VIS community.

\textit{2) Expansion to the RGB-D scenarios:}
To investigate the adaptability of our proposed method to other modalities, we apply it to the RGB-D semantic segmentation dataset NYUDepthV2 \cite{Silberman:ECCV12}, as shown in Table \ref {tab:rgbd}.
Notably, without any modifications, we feed depth images into the infrared input stream.
Under such circumstances, our method still outperforms the full fine-tuning paradigm. 
Moreover, leveraging the powerful capabilities of EVA02 \cite{fang2024eva}, the performance of IV-tuning outperforms the previous state-of-the-art methods, which demonstrates the effectiveness of our proposed method. 
It should be noted that although IV-tuning remains effective in the RGB-D domain, for the depth modality, modality-specific designs may be required to achieve better tuning performance. 

\subsection{Analysis of Overfitting}
Extensive experiments validate IV-tuning's effectiveness in adapting PVMs to downstream IR-VIS tasks.
We hypothesize that the success of such adaptation may stem from IV-tuning mitigating overfitting of PVMs on small datasets, resulting in improved generalization.
To verify this, we calculate the average training loss of the last 1,000 iterations for semantic segmentation and that of the last one epoch for object detection, respectively, as well as their corresponding evaluation metrics.
As shown in Fig. \ref{fig:loss and metric}, as the training parameters increase from Freeze to IV-tuning to full fine-tuning (Full), the training loss monotonically decreases, suggesting that more trainable parameters can make the model better fit on the training set.
However, the test metric first increases then decreases, indicating that the full fine-tuning paradigm overfit on the training set.
In contrast, IV-tuning achieves a better trade-off: it improves fitting capacity compared to the Freeze baseline while effectively avoiding the overfitting trap of full fine-tuning.
This observation aligns with the conclusion in Rein \cite{wei2024stronger} and Fig. \ref{fig:pca}, providing strong evidence for the limitations of full fine-tuning discussed earlier and supporting our motivation to streamline the infrared flow and leverage PVMs with fewer trainable parameters for robust generalization.
Moreover, this phenomenon also applies to the comparison of Rein \cite{wei2024stronger} and IV-tuning, indicating that our IV-tuning better balances generalization with the degree of training.

\section{Conclusion}
\label{conclusion}
This paper investigates the critical generalization bottleneck in existing IR-VIS methods, revealing that the dual-branch full fine-tuning paradigm leads to a highly constrained and low-ranked feature space.
To address this, we propose IV-tuning, a streamlined parameter-efficient transfer learning framework that effectively harnesses the power of modern pre-trained visual models for multi-modal tasks.
Grounded in our frequency spectrum and intrinsic dimensionality analysis, we introduce the Modality-aware Prompter, which incorporates two key innovations: (1) a modality-specific design that employs simple linear projections to preserve the crucial low-frequency thermal priors of the infrared modality, and (2) a rank-adaptive fusion strategy that aligns fusion dimensions with the feature complexity at different depths.
Extensive experiments across three downstream tasks, including salient object detection, semantic segmentation, and object detection, demonstrate that IV-tuning not only significantly reduces computational overhead, but also achieves superior generalization performance compared to state-of-the-art infrared-visible and parameter-efficient transfer learning methods.
We hope this work provides a new perspective on efficiently adapting large-scale foundation models for the broader multi-modal community.

% \section*{Acknowledgments}
% This should be a simple paragraph before the References to thank those individuals and institutions who have supported your work on this article.

\bibliographystyle{IEEEtran}
\bibliography{IEEEabrv,mylib}

\par\noindent 
\parbox[t]{\linewidth}{
\noindent\parpic{\includegraphics[height=1.5in,width=1in,clip,keepaspectratio]{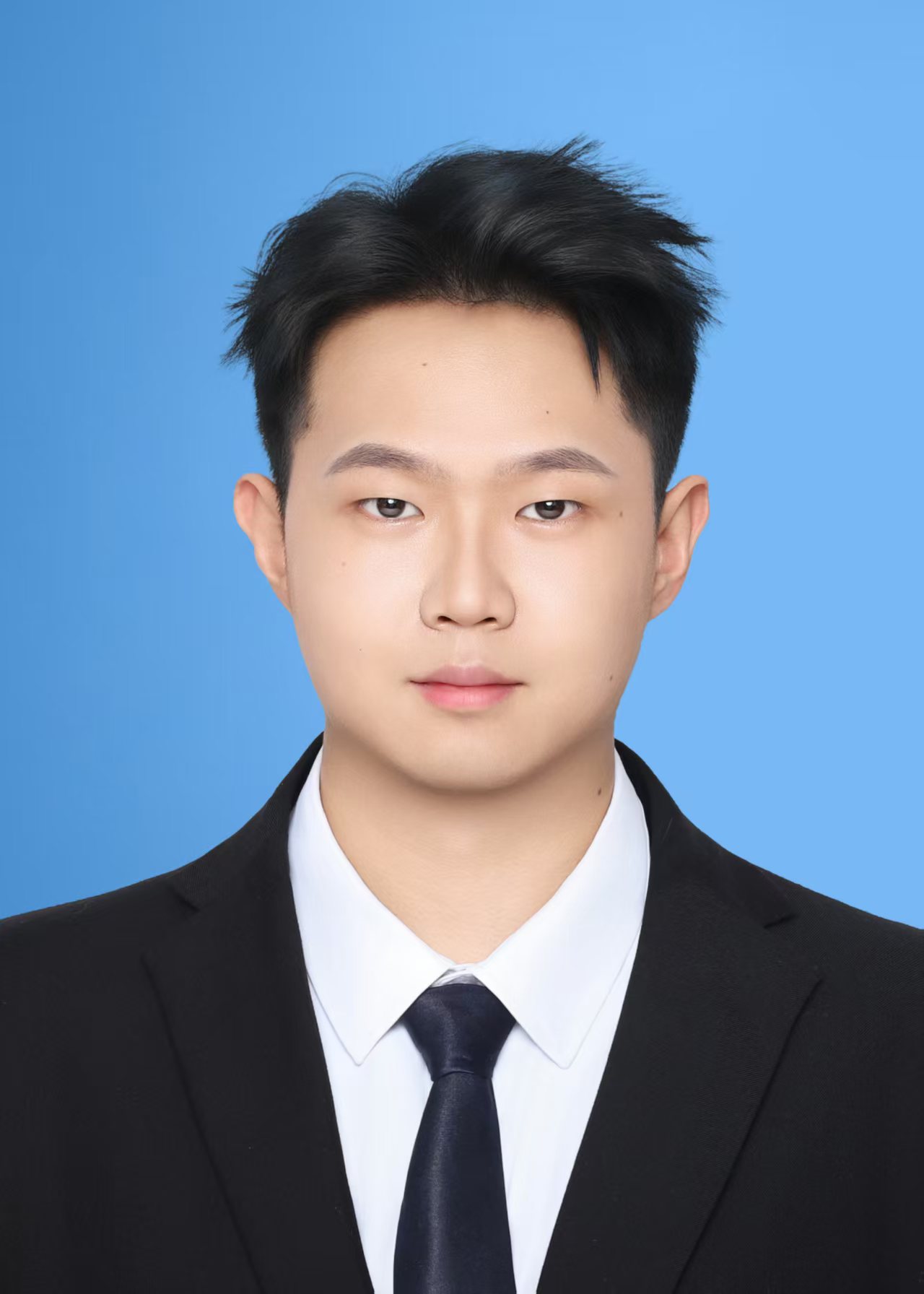}}
\noindent {\bf Yaming Zhang}\
received the B.S. from the School of Communications and Information Engineering at Chongqing University of Posts and Telecommunications in 2023. He is currently pursuing an academic Master's degree in the same college and university. His research interests include efficient fine-tuning of deep learning and multi-modal learning.}
\vspace{0.2\baselineskip}

\par\noindent 
\parbox[t]{\linewidth}{
\noindent\parpic{\includegraphics[height=1.5in,width=1in,clip,keepaspectratio]{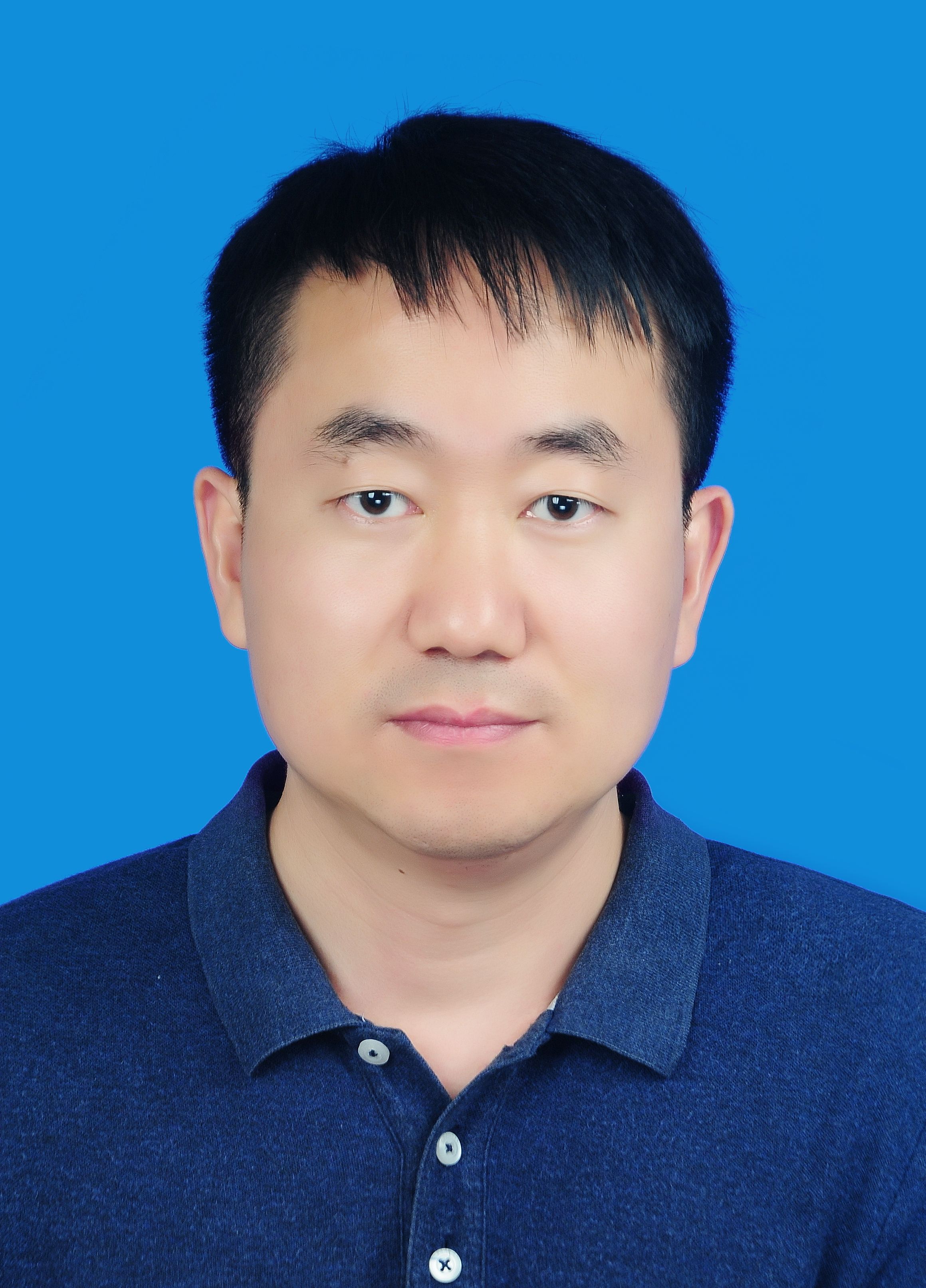}}
\noindent {\bf Chenqiang Gao}\
received the B.S. degree in computer science from the China University of Geosciences, Wuhan, China, in 2004 and the Ph.D. degree in control science and engineering from the Huazhong University of Science and Technology, Wuhan, China, in 2009. In August 2009, he joined the School of Communications and Information Engineering, Chongqing University of Posts and Telecommunications (CQUPT), Chongqing, China. In September 2012, he joined the Informedia Group with the School of Computer Science, Carnegie Mellon University, Pittsburgh, PA, USA, working on multimedia event detection (MED) and surveillance event detection (SED) until March 2014, when he returned to CQUPT. In September 2023, he joined the School of Intelligent Systems Engineering, Sun Yat-sen University, Shenzhen, Guangdong, China. His research interests include image processing, infrared target detection, action recognition, and event detection.}
\vspace{0.2\baselineskip}

\par\noindent 
\parbox[t]{\linewidth}{
\noindent\parpic{\includegraphics[height=1.5in,width=1in,clip,keepaspectratio]{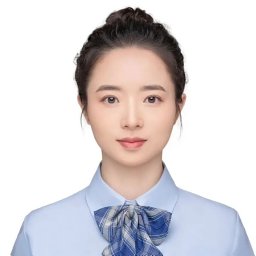}}
\noindent {\bf Fangcen Liu}\
received the B.S., M.S., and Ph.D. degrees from the School of Communications and Information Engineering, Chongqing University of
Posts and Telecommunications, Chongqing, China, in 2018, 2021, and 2025, respectively. In 2025, she joined the School of Computer Science and Technology, Hainan University, Haikou, China, where she is currently an Associate Professor. Her research interests include infrared small target detection, foundation model, and infrared-visible vision task understating.
}
\vspace{0.2\baselineskip}

\par\noindent 
\parbox[t]{\linewidth}{
\noindent\parpic{\includegraphics[height=1.5in,width=1in,clip,keepaspectratio]{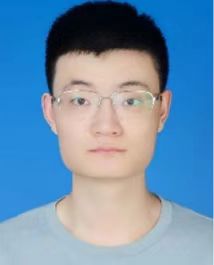}}
\noindent {\bf Junjie Guo}\
received the B.S. and M.S. degrees from the School of Communications and information Engineering, Chongqing University of Posts and Telecommunications, Chongqing, China, in 2022 and 2025, respectively. 
}
\vspace{0.2\baselineskip}

\par\noindent 
\parbox[t]{\linewidth}{
\noindent\parpic{\includegraphics[height=1.5in,width=1in,clip,keepaspectratio]{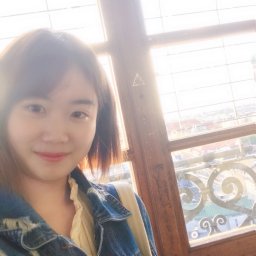}}
\noindent {\bf Lan Wang}\
received the M.S. degree from the School of Communications and Information Engineering, Chongqing University of Posts and Telecommunications, Chongqing, China. She is currently pursuing the Ph.D. degree at Michigan State University, East Lansing, MI, USA.
}
\vspace{0.2\baselineskip}

\par\noindent 
\parbox[t]{\linewidth}{
\noindent\parpic{\includegraphics[height=1.5in,width=1in,clip,keepaspectratio]{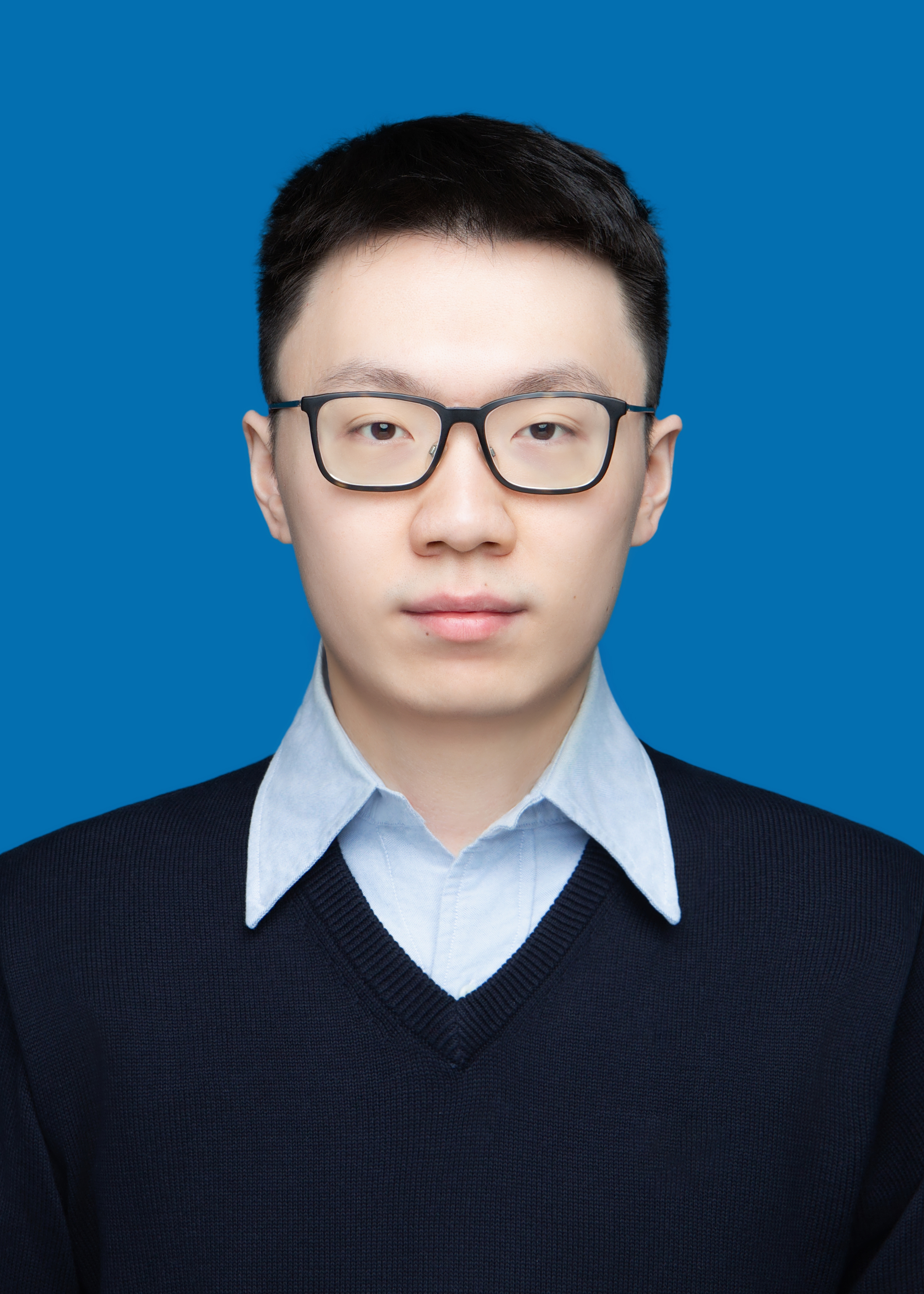}}
\noindent {\bf Xinggan Peng}\
received the Ph.D. degree t school of Electrical and Electronic Engineering, Nanyang Technological University, in 2024. He received the B.Eng. (Hons.) degree from both University of Electronic Science and Technology of
China and University of Glasgow in 2017. He received the M.S. degree in Electrical and Computer Engineering 
from The Ohio State University in 2019. He is currently the post-doc at CMCU Engineering Co,.Ltd. His main interests include signal processing, intelligent transportation system and deep learning.
}
\vspace{0.2\baselineskip}

\par\noindent 
\parbox[t]{\linewidth}{
\noindent\parpic{\includegraphics[height=1.5in,width=1in,clip,keepaspectratio]{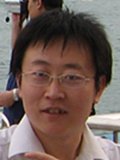}}
\noindent {\bf Deyu Meng}\
(Member, IEEE) received the B.Sc., M.Sc., and Ph.D. degrees from Xi’an Jiaotong University, Xi’an, China, in 2001, 2004, and 2008, respectively. He is currently a Professor with the Institute for Information and System Sciences, Xi’an Jiaotong University. He was a Visiting Scholar with Carnegie Mellon University, Pittsburgh, PA, USA, from 2012 to 2014. His current research interests include self-paced learning, noise modeling, and tensor sparsity.
}
\vspace{0.2\baselineskip}

\newpage

% \section{Biography Section}
% If you have an EPS/PDF photo (graphicx package needed), extra braces are
%  needed around the contents of the optional argument to biography to prevent
%  the LaTeX parser from getting confused when it sees the complicated
%  $\backslash${\tt{includegraphics}} command within an optional argument. (You can create
%  your own custom macro containing the $\backslash${\tt{includegraphics}} command to make things
%  simpler here.)
 
% \vspace{11pt}

% \bf{If you include a photo:}\vspace{-33pt}
% \begin{IEEEbiography}[{\includegraphics[width=1in,height=1.25in,clip,keepaspectratio]{fig1}}]{Michael Shell}
% Use $\backslash${\tt{begin\{IEEEbiography\}}} and then for the 1st argument use $\backslash${\tt{includegraphics}} to declare and link the author photo.
% Use the author name as the 3rd argument followed by the biography text.
% \end{IEEEbiography}

% \vspace{11pt}

% \bf{If you will not include a photo:}\vspace{-33pt}
% \begin{IEEEbiographynophoto}{John Doe}
% Use $\backslash${\tt{begin\{IEEEbiographynophoto\}}} and the author name as the argument followed by the biography text.
% \end{IEEEbiographynophoto}

\vfill

\end{document}